\def\paperlanguage{en} 

\documentclass[letterpaper, 10 pt, conference]{ieeeconf}  

\usepackage[dvips]{graphicx}
\usepackage{bm}
\usepackage{cite}
\usepackage{flushend}
\include{preamble}
\usepackage{booktabs}

\IEEEoverridecommandlockouts                              

\title{\LARGE \textbf
  {
    \switchlanguage%
    {%
			EFGCL: Learning Dynamic Motion through Spotting-Inspired External Force Guided Curriculum Learning
		}%
    {%
			EFGCL: Learning Dynamic Motion through Spotting-Inspired External Force Guided Curriculum Learning
    }%
  }
}

\author{Keita Yoneda$^{1}$, Kento Kawaharazuka$^{1, 2}$, Kei Okada$^{1}$
  \thanks{$^{1}$ The authors are with the Department of Mechano-Informatics, Graduate School of Information Science and Technology, The University of Tokyo, 7-3-1 Hongo, Bunkyo-ku, Tokyo, 113-8656, Japan.
    {\texttt\small [yoneda, kawaharazuka, k-okada]@jsk.imi.i.u-tokyo.ac.jp} 
  }
  \thanks{$^{2}$ The author is with the AI Center, Graduate School of Information Science and Technology, The University of Tokyo, Japan.
  }
} 
\begin{document}

\maketitle
\thispagestyle{empty}
\pagestyle{empty}

\begin{abstract}
\switchtext%
{
Learning dynamic whole-body motions for legged robots through reinforcement learning (RL) remains challenging due to the high risk of failure, which makes efficient exploration difficult and often leads to unstable learning.

In this paper, we propose External Force Guided Curriculum Learning (EFGCL), a guided RL approach based on the principle of \emph{physical guidance}, in which external assistive forces are introduced during training.
Inspired by spotting in artistic gymnastics, EFGCL enables agents to physically experience successful motion executions without relying on task-specific reward shaping or reference trajectories.

Experiments on a quadrupedal robot performing Jump, Backflip, and Lateral-Flip tasks demonstrate that EFGCL accelerates learning of the Jump task by approximately a factor of two and enables the acquisition of complex whole-body motions that conventional RL methods fail to learn.
We further show that the learned policies can be deployed on a real robot, reproducing motions consistent with those observed in simulation.

These results indicate that physically guided exploration, which allows agents to experience success early in training, is an effective and general strategy for improving learning efficiency in dynamic whole-body motion tasks.
}
{
動的な全身運動を脚式ロボットに強化学習(Reinforcement Learning; RL)によって獲得させることは,失敗リスクが高く探索が困難であるため,依然として大きな課題である.

本研究では,学習過程において外部補助力を導入することで探索を誘導する,Physical Guidance の原理に基づく Guided-RL 手法として,External Force Guided Curriculum Learning(EFGCL)を提案する.
EFGCL は体操競技における Spotting に着想を得たものであり,タスク固有の報酬設計や参照軌道に依存することなく,エージェントに成功動作を物理的に体験させる枠組みを提供する.

四脚ロボットによる Jump,Backflip,Lateral-Flip タスクを用いた実験により,EFGCL は Jump タスクの学習を約 2 倍高速化するとともに,従来の RL 手法では学習が困難であった複雑な全身動作の獲得を可能にすることを示した.
さらに,学習された方策を実機ロボットに展開し,シミュレーションと整合した動作が再現可能であることを確認した.

これらの結果は,成功動作を物理的に体験させるという単純な原理が,動的全身運動における学習効率を向上させ,強化学習の適用範囲を拡張する有効なアプローチであることを示している.
}
{
}
\end{abstract}

\section{Introduction}\label{sec:introduction}
\switchtext%
{%
To achieve high locomotion performance in unstructured environments, quadrupedal robots require learning methods that can stably acquire a wide range of diverse and complex motor skills.
With recent advances in reinforcement learning (RL), numerous approaches have been proposed that enable robust learning of individual behaviors, such as locomotion over rough terrain \cite{hwangbo2019anymal, miki2022anymal, zhuang2023robot, vogel2024robust_ladder, kim2025high}.

For tasks such as rough-terrain locomotion and obstacle traversal, learning methods with high success rates have already been established.
In contrast, learning dynamic motor skills involving high acceleration and high energy, as exemplified by sports motions \cite{yoneda2025kleiyn}, still requires substantial task-specific tuning.
This difficulty arises because motions with a high risk of failure are inherently difficult to explore, and learning rarely progresses without explicit guidance.

To address the learning of such dynamic motions, Guided Reinforcement Learning (Guided-RL), which introduces assistance during the learning process, has been widely studied \cite{esser2022guided}.
Representative approaches include imitation learning based on reference trajectories and reward shaping, which guides behavior through carefully designed reward functions.

\begin{figure}[t]
    \centering
        \includegraphics[width=0.95\columnwidth]{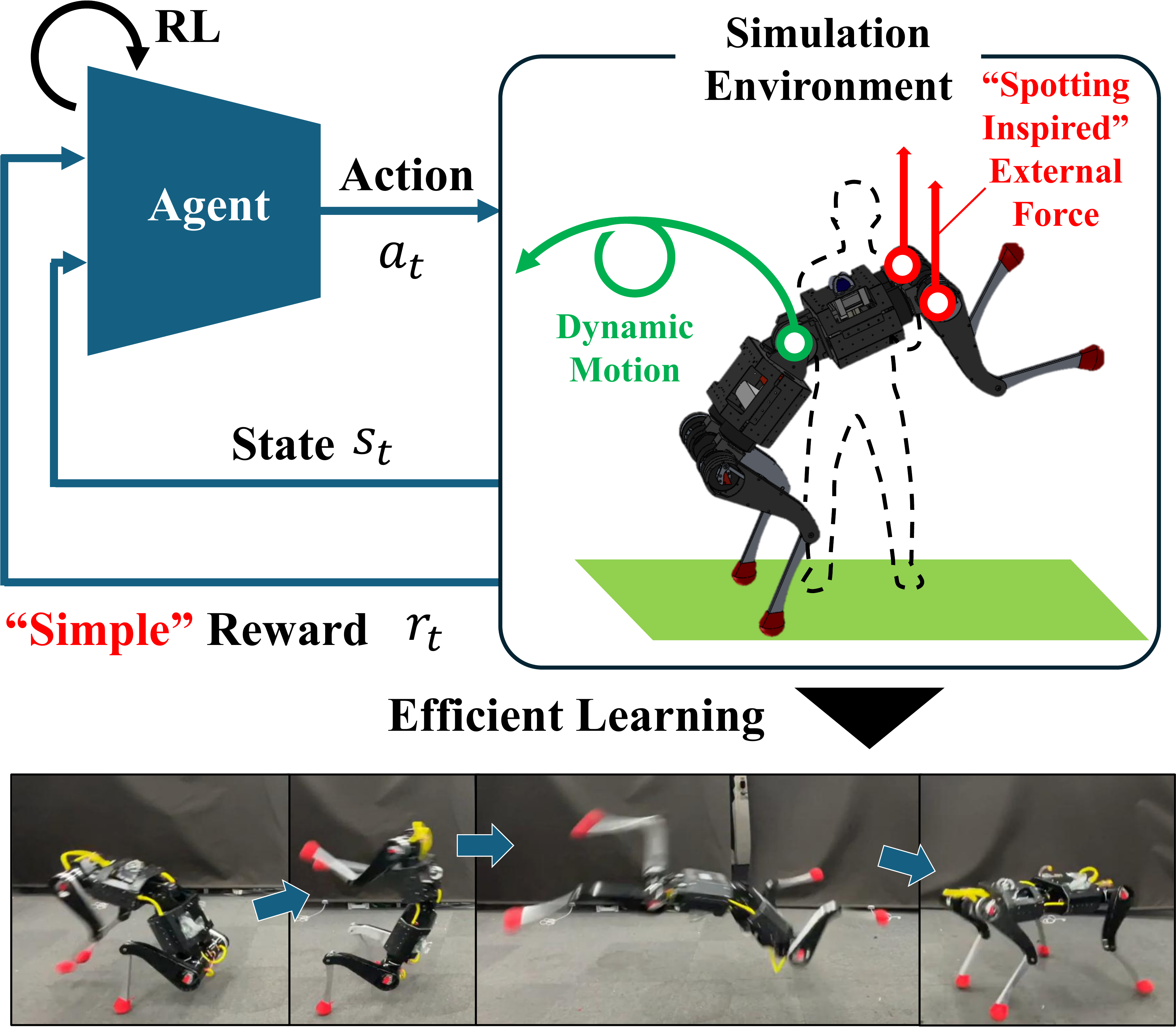}
        \caption{Conceptual overview of External Force Guided Curriculum Learning (EFGCL).
        By applying external assistive forces in the early stages of learning, the agent experiences motion sequences with a higher probability of success.
        As learning progresses, the assistance is gradually reduced in a curriculum manner, ultimately enabling the agent to acquire a policy that achieves the target motion without assistance.}
    \label{fig:efgcl_overview}
        \vspace{-3ex}
\end{figure}

Imitation learning directly mimics reference trajectories that represent target motions and has been increasingly applied to dynamic tasks \cite{peng2018deepmimic, peng2020learning, wu2023learning}.
However, the performance of the learned motions strongly depends on the quality of the reference trajectories.
Methods such as Opt-Mimic \cite{fuchioka2022opt}, which generate trajectories through optimization, can provide high-quality references, but they incur substantial costs in robot modeling and objective function design.
Alternatively, approaches such as WASABI \cite{li2023learning} utilize demonstrations obtained by physically guiding the robot, reducing data collection costs.
However, the quality of such data depends heavily on the skill of the human operator, making it difficult to ensure stability.
As a result, there exists an inherent trade-off between the quality of reference trajectories and the cost of generating them, and achieving high performance solely through imitation learning remains expensive.

Reward shaping aims to facilitate exploration by designing intermediate rewards that capture key elements of the desired behavior \cite{devlin2011theoretical, atanassov2024curriculum, bellegarda2024robust}.
However, determining which aspects of a motion should be defined as intermediate rewards is highly task-dependent and non-trivial.
Moreover, intermediate rewards may introduce designer bias, which can exclude potentially optimal motion sequences and degrade learning performance \cite{vasan2024revisiting}.
For these reasons, \cite{vasan2024revisiting} recommends using sparse reward functions and promoting learning through design choices outside the reward itself.
Nevertheless, a general framework for efficiently exploring dynamic motions with a high risk of failure has yet to be established.

Insightful inspiration can be drawn from artistic gymnastics, where dynamic motions are the primary objective.
In gymnastics training, a technique known as \emph{spotting} is commonly used, in which a coach physically supports the athlete while practicing a skill \cite{sorzano2023spotting}.
This approach assists the exploration process through physical guidance and differs fundamentally from conventional methods that guide behavior solely through reward design.

Motivated by this idea, we propose \emph{External Force Guided Curriculum Learning} (EFGCL), which applies this principle to reinforcement learning for robots (\figref{fig:efgcl_overview}).
The main contributions of this work are summarized as follows:
\begin{itemize}
    \item We introduce a new learning paradigm for dynamic motor skill acquisition that employs \emph{physical guidance via external forces}, rather than guidance through reward design.
    \item We propose External Force Guided Curriculum Learning (EFGCL), which gradually decays external assistive forces and demonstrate that physical guidance significantly improves exploration efficiency.
    \item Through learning experiments on a real quadrupedal robot, we demonstrate that the proposed approach is effective and transferable to real-world environments.
\end{itemize}
}
{%
四脚ロボットが不整地環境において高い移動性能を発揮するためには,多様で複雑な運動スキルを安定して獲得できる学習手法が不可欠である.
近年の強化学習(Reinforcement Learning; RL)の発展により,不整地歩行などの個々の動作を高いロバスト性で学習する手法が数多く提案されてきた\cite{hwangbo2019anymal, miki2022anymal, zhuang2023robot, vogel2024robust_ladder, kim2025high}.

不整地歩行や障害物踏破といったタスクに対しては,高い成功率を示す学習手法がすでに確立されつつある.
一方で, スポーツ動作\cite{yoneda2025kleiyn}に代表される高加速度・高エネルギーを伴う動的な運動スキルの学習は, 依然としてタスク固有のチューニングを必要とする.
これは, 失敗リスクの高い動作は探索が難しく, 誘導なしでは学習が進行しにくいためである.

このような動的動作の学習に対しては,従来より学習過程に補助を導入するGuided-RL\cite{esser2022guided} が用いられてきた.
代表的なアプローチとしては,参照軌道を用いる模倣学習や,報酬設計により動作を誘導する Reward Shaping が挙げられる.

\begin{figure}[t]
    \centering
        \includegraphics[width=0.95\columnwidth]{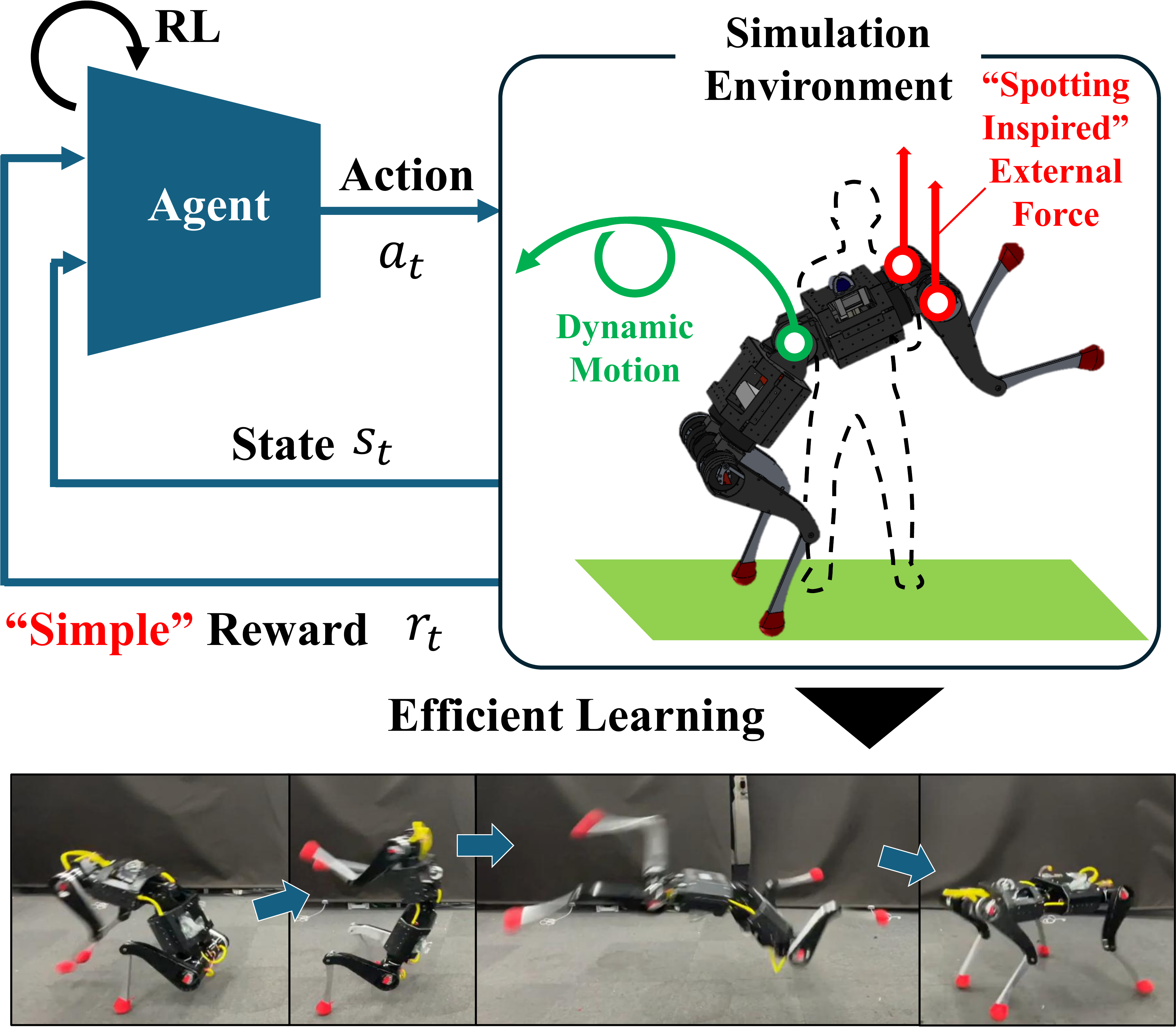}
        \caption{External Force Guided Curriculum Learning (EFGCL) の概念図. 学習初期に外部補助力を付与することで成功しやすい運動系列を経験させ, 学習の進行に伴って補助力を段階的に減衰させることで, 最終的に補助なしで目標動作を達成する方策を獲得する.}
    \label{fig:efgcl_overview}
		\vspace{-3ex}
\end{figure}

模倣学習は,目標動作を表す参照軌道を直接模倣する手法であり,動的タスクへの適用も進んでいる\cite{peng2018deepmimic, peng2020learning, wu2023learning}.
しかし,学習される動作の性能は参照軌道の品質に強く依存する.
Opt-Mimic\cite{fuchioka2022opt} のように軌道を最適化により生成する手法は,高品質な参照軌道を提供できる.
しかしロボットのモデル作成や目的関数の作成に大きなコストを要するという欠点がある.
WASABI\cite{li2023learning} のように人の手でロボットを動かしたデータを用いる場合にはデータの作成コストを抑えられる.
しかしその反面データ品質が操作者の熟練度に依存し,安定性を確保することが難しくなる.
ここから,参照軌道の品質と作成コストの間には本質的なトレードオフが存在し, 模倣学習のみで高い性能を得ることは依然として高コストであると言える.

Reward Shaping は,動作の重要要素を中間報酬として設計することで探索を容易にすることを目的とした手法である\cite{devlin2011theoretical, atanassov2024curriculum, bellegarda2024robust}.
しかし,どのような動作を中間報酬として定義すべきかはタスクに強く依存し,非自明である.
また,中間報酬には設計者のバイアスが含まれる可能性がある.
バイアスは本来存在し得る最適な運動系列を排除して学習性能を低下させる可能性が指摘されている\cite{vasan2024revisiting}.
こうした理由から\cite{vasan2024revisiting} では疎な報酬関数を用いて, 報酬以外の部分の設計によって学習を促進することが推奨されている.
しかし動的な動作で失敗リスクの高い動作の探索を効率化するための一般的な枠組みは未だ確立されていない.

ここで動的な動作の実現を目的とする体操競技に注目すると興味深い示唆が得られる.
体操競技では技の習得時に「Spotting」という, 選手を外部から人が支えつつ技を練習する方法が用いられる\cite{sorzano2023spotting}.
これは物理的な補助によって探索過程を支援するものであり, 報酬設計を通じて行動を誘導する従来手法とは異なる枠組みである.

本研究では, この発想をロボットの強化学習に応用した External Force Guided Curriculum Learning(EFGCL)を提案する(\figref{fig:efgcl_overview}).
本研究の主な Contribution は以下のとおりである.
\begin{itemize}
    \item 動的運動スキルの学習において, 報酬設計による誘導ではなく, 外力による Physical Guidance を用いるという新しい学習パラダイムを提示した.
    \item 外部補助力を段階的に減衰させる External Force Guided Curriculum Learning (EFGCL) を提案し, Physical Guidance によって探索が効率化されることを示した.
    \item 実機四脚ロボットを用いた学習実験を通じて, 提案手法が現実環境においても有効に機能することを実証した.
\end{itemize}
}%
{%
}%

\section{Background}\label{sec:background}
\switchtext
{
\subsection{Proximal Policy Optimization (PPO)}\label{sec:ppo}
In reinforcement learning for legged robots, Proximal Policy Optimization (PPO) \cite{schulman2017ppo} is commonly used due to its training stability and ease of implementation.
PPO constrains the magnitude of policy updates by optimizing a clipped surrogate objective function based on the likelihood ratio between the current and previous policies,
\[
r_t(\theta)=\frac{\pi_\theta(a_t\mid s_t)}{\pi_{\theta_{\text{old}}}(a_t\mid s_t)}, \quad a_t\sim\pi_{\theta_{\text{old}}}(\cdot\mid s_t),
\]
thereby enabling stable learning while limiting excessive policy updates.
This update control is particularly effective for legged robots with high-dimensional action spaces and unstable dynamics, as it prevents training collapse caused by abrupt policy changes.

PPO also employs Generalized Advantage Estimation (GAE) \cite{schulman2015high} to estimate the advantage $A_t$.
GAE computes the advantage as a weighted sum of temporal-difference (TD) errors,
$\delta_t = r_t + \gamma V(s_{t+1}) - V(s_t)$,
where $\gamma \in [0,1]$ is the discount factor.
This formulation allows a trade-off between variance and bias.
Based on the estimated $A_t$, PPO updates the policy in the following gradient direction:
\[
\nabla_\theta J(\theta)
\propto
\mathbb{E}_{s_t,a_t\sim\pi_\theta}
\left[
\nabla_\theta \log \pi_\theta(a_t\mid s_t) A_t
\right].
\]
However, in environments with a high risk of failure, unsuccessful trajectories tend to dominate, resulting in small estimated state values $V(s_t)$ for many states.
Consequently, both the TD error $\delta_t$ and the advantage $A_t$ approach zero, providing little useful information for gradient-based updates.
Thus, while PPO is a stable optimization method, learning can stagnate severely in environments where successful experiences are rarely obtained.

\begin{algorithm}[t]
\caption{External Force Guided Curriculum Learning}
\label{alg:efgcl}
\begin{algorithmic}[1]
\State \textbf{Input:} initial policy $\pi_{\text{init}}$, initial critic $V_{\text{init}}$, assist force $F_{\text{assist}}$, decay step size $\varepsilon$, success threshold $\zeta$
\State \textbf{Output:} final policy $\pi_N$

\State Initialize:
\[
F_0 \leftarrow F_{\text{assist}},\quad
\pi_0 \leftarrow \pi_{\text{init}},\quad
V_0 \leftarrow V_{\text{init}}
\]

\State Set decay rate:
\[
\alpha \leftarrow 1.0
\]

\For{$i = 0, \dots, N-1$}
    \State $success\_rate \leftarrow 0$
    \While{$success\_rate < \zeta$}
        \State Train $\pi_i$ and $V_i$ using PPO under assist force $F_i$
        \State Update $success\_rate$
    \EndWhile

    \State Decay assist force:
    \[
    \alpha \leftarrow \max(0, 1 - \varepsilon \times i),\quad
    F_{i+1} \leftarrow \alpha \times F_{\text{assist}}
    \]

    \State Carry over the learned policy and value function:
    \[
    \pi_{i+1} \leftarrow \pi_i,\quad
    V_{i+1} \leftarrow V_i
    \]
\EndFor

\State \Return $\pi_N$
\end{algorithmic}
\end{algorithm}

\subsection{Curriculum Learning}\label{sec:curriculum_learning}
Curriculum learning has been widely adopted as a representative approach to address the aforementioned issue.

In curriculum learning, training begins in specialized environments with low exploration risk, and the task difficulty or risk level is gradually increased.
This framework can be interpreted as sequential learning over multiple Markov Decision Processes (MDPs) $\mathcal{M}_i$ with different risk levels.
If the changes in dynamics and reward structures between successive stages are sufficiently small, the optimal policy $\pi_i^*$ for stage $\mathcal{M}_i$ and the optimal policy $\pi_{i+1}^*$ for the subsequent stage $\mathcal{M}_{i+1}$ are expected to be similar.
That is,
\[
\frac{\pi_{i+1}^*(a_t\mid s_t)}{\pi_i^*(a_t\mid s_t)} \approx 1, \quad a_t\sim\pi_i^*(\cdot\mid s_t),
\]
can be artificially constructed.
This property is consistent with the assumption underlying PPO that smaller policy update steps are preferable.
Therefore, curriculum learning provides a theoretically well-aligned framework that supports stable, incremental learning with PPO.

\subsection{Stabilizing Learning via Accelerated Value Function Estimation}\label{sec:rapid_value_estimation}
In reinforcement learning with sparse rewards, learning efficiency strongly depends on how quickly the value function can correctly evaluate action sequences that yield high rewards.
This is particularly critical for methods such as PPO, which rely on value functions for advantage estimation, where the initial accuracy of value estimation significantly influences the direction of policy updates.

In general, once a sufficient number of high-reward trajectories are observed, the state values $V(s)$ corresponding to the associated state sequence
$ \xi = \{ s_t^{\text{high}} \}_{t=0}^{N} $
are estimated to be large.
When such high-value trajectories exist, actions that deviate from them yield large negative TD errors and advantages.
As a result, the policy is updated in a direction that discourages deviation from high-reward trajectories, making effective motion sequences more likely to be preserved once acquired.

Therefore, whether the value function can assign high values to near-optimal behaviors from the early stages of learning is a crucial factor that determines the overall stability and efficiency of training.
To achieve this, it is effective to expose the agent to a large number of high-reward trajectories during the early phase of learning.

}
{
\subsection{Proximal Policy Optimization (PPO)}\label{sec:ppo}
脚ロボットの強化学習では,学習の安定性と実装容易性の観点から Proximal Policy Optimization (PPO) \cite{schulman2017ppo} が標準的に用いられている.
PPO は過去方策との尤度比
\[
r_t(\theta)=\frac{\pi_\theta(a_t\mid s_t)}{\pi_{\theta_{\text{old}}}(a_t\mid s_t)}, \quad a_t\sim\pi_{\theta_{\text{old}}}(\cdot\mid s_t)
\]
を用いたクリップ付き代理目的関数を最適化することで,方策更新幅を制限しながら学習を進める.
この更新制御は,高次元アクション空間や不安定なダイナミクスを有する脚ロボットにおいて特に有効であり,急激な方策変化による学習崩壊を防ぐ役割を果たす.

また PPO では,Generalized Advantage Estimation (GAE) \cite{schulman2015high} を用いてアドバンテージ $A_t$ を推定する.
GAE は TD 誤差 $ \delta_t = r_t + \gamma V(s_{t+1}) - V(s_t) $ の重み付き和としてアドバンテージを計算し,分散とバイアスのトレードオフを調整する. ここで$\gamma$は割引率である.
PPO は推定された $A_t$ に基づき,以下の勾配方向に方策を更新する.
\[
\nabla_\theta J(\theta)
\propto
\mathbb{E}_{s_t,a_t\sim\pi_\theta}
\left[
\nabla_\theta \log \pi_\theta(a_t\mid s_t) A_t
\right]
\]
しかし一般に,失敗リスクの高い環境では失敗軌道が多数を占めるため,多くの状態 $s_t$ に対して状態価値 $V(s_t)$ は小さく推定される.
この結果,TD 誤差 $\delta_t$ およびアドバンテージ $A_t$ は 0 近傍の値となり,勾配更新に利用可能な情報がほとんど得られなくなる.
したがって PPO は安定な最適化手法である一方で,成功経験が得られにくい環境では学習が著しく停滞しやすいという課題を抱えている.

\subsection{Curriculum Learning}\label{sec:curriculum_learning}
前節の問題に対する代表的な対処法として,カリキュラム学習が用いられてきた.

カリキュラム学習では,探索時のリスクが低い特殊な環境から学習を開始し,段階的にタスクの難易度やリスクを高めていく.
この枠組みは,リスクレベルの異なる複数の Markov Decision Process (MDP) $\mathcal{M}_i$ 上での逐次的な学習として捉えることができる.
各ステージ間でダイナミクスや報酬構造の変化が十分に小さい場合,ステージ $\mathcal{M}_i$ における最適方策 $\pi_i^*$ と,次のステージ $\mathcal{M}_{i+1}$ における最適方策 $\pi_{i+1}^*$ の差異も小さいと期待される. すなわち
\[
\frac{\pi_{i+1}^*(a_t\mid s_t)}{\pi_i^*(a_t\mid s_t)} \approx 1, \quad a_t\sim\pi_i^*(\cdot\mid s_t)
\]
が成り立つ状況を人工的に構成していると言え, これは PPO が前提とする「更新幅は小さいほど望ましい」という仮定と整合的である.
よってカリキュラム学習は PPO による段階的な学習の安定性を理論的にも支持する枠組みであると言える.

\subsection{価値関数の推定高速化による学習安定化}\label{sec:rapid_value_estimation}
疎な報酬を用いた強化学習における学習効率は,高い報酬を得られる行動系列を価値関数がどれだけ早期に正しく評価できるかに大きく依存する.
特に PPO のようにアドバンテージ推定に価値関数を用いる手法では,価値推定の初期精度が方策更新の方向性を大きく左右する.

一般に高い報酬を得られる軌道が一定数観測されると,それに対応する状態列 $ \xi = \{ s_t^{\text{high}} \}_{t=0}^{N} $ に対して状態価値 $V(s)$ は大きく推定される.
そしてこのような高価値軌道が存在する場合,そこから逸脱する行動に対しては TD 誤差およびアドバンテージが大きな負の値として計算される.
その結果,方策は高報酬軌道から外れない方向へと更新され,一度獲得された有効な運動系列が維持されやすくなる.

以上より,価値関数が学習初期から正解に近い動作を高く評価できるかどうかが,学習全体の安定性と効率を決定づける重要な要因であると言え, そのためには学習初期に高報酬軌道を多く経験させることが有効である.

\begin{algorithm}[t]
\caption{External Force Guided Curriculum Learning}
\label{alg:efgcl}
\begin{algorithmic}[1]
\State \textbf{Input:} initial policy $\pi_{\text{init}}$, initial critic $V_{\text{init}}$, 
assist force $F_{\text{assist}}$, decay step size $\varepsilon$, success threshold $\zeta$
\State \textbf{Output:} final policy $\pi_N$

\State Initialize:
\[
F_0 \leftarrow F_{\text{assist}},\quad
\pi_0 \leftarrow \pi_{\text{init}},\quad
V_0 \leftarrow V_{\text{init}}
\]

\State Set decay rate:
\[
	\alpha \leftarrow 1.0
\]

\For{$i = 0, \dots, N-1$}
    \State $success\_rate \leftarrow 0$
    \While{$success\_rate < \zeta$}
        \State Train $\pi_i$ and $V_i$ using PPO under assist force $F_i$
        \State Update $success\_rate$
    \EndWhile

    \State Decay assist force:
    \[
    \alpha \leftarrow \max(0, 1 - \varepsilon \times i),\quad
    F_{i+1} \leftarrow \alpha \times F_{\text{assist}}
    \]

    \State Carry over the learned policy and value function:
    \[
    \pi_{i+1} \leftarrow \pi_i,\quad
    V_{i+1} \leftarrow V_i
    \]
\EndFor

\State \Return $\pi_N$
\end{algorithmic}
\end{algorithm}
}
{
}

\section{Method}\label{sec:method}
\switchtext
{
\subsection{Overview of External Force Guided Curriculum Learning (EFGCL)}\label{sec:efgcl_overview}
The proposed External Force Guided Curriculum Learning (EFGCL) provides a framework for achieving stable learning in dynamic tasks while maintaining sparse reward functions.
Instead of modifying rewards as in imitation learning or Reward Shaping, EFGCL stabilizes learning by curriculum-wise modifying the Markov Decision Process (MDP) itself in which learning is performed.

The overall procedure of the algorithm is summarized in \algoref{alg:efgcl}.
EFGCL designs external assistive forces that facilitate task execution during the early stage of learning (line~1),
and gradually decays this assistance as training progresses (lines~7--9).
Through this curriculum process, the agent experiences trajectories with a high probability of success under assistance,
while gradually transitioning toward autonomous motion generation.
Eventually, the agent acquires a policy $\pi_N$ that achieves the target motion without assistance (line~14).

\begin{figure}[tbp]
\centering
	\includegraphics[width=0.9\columnwidth]{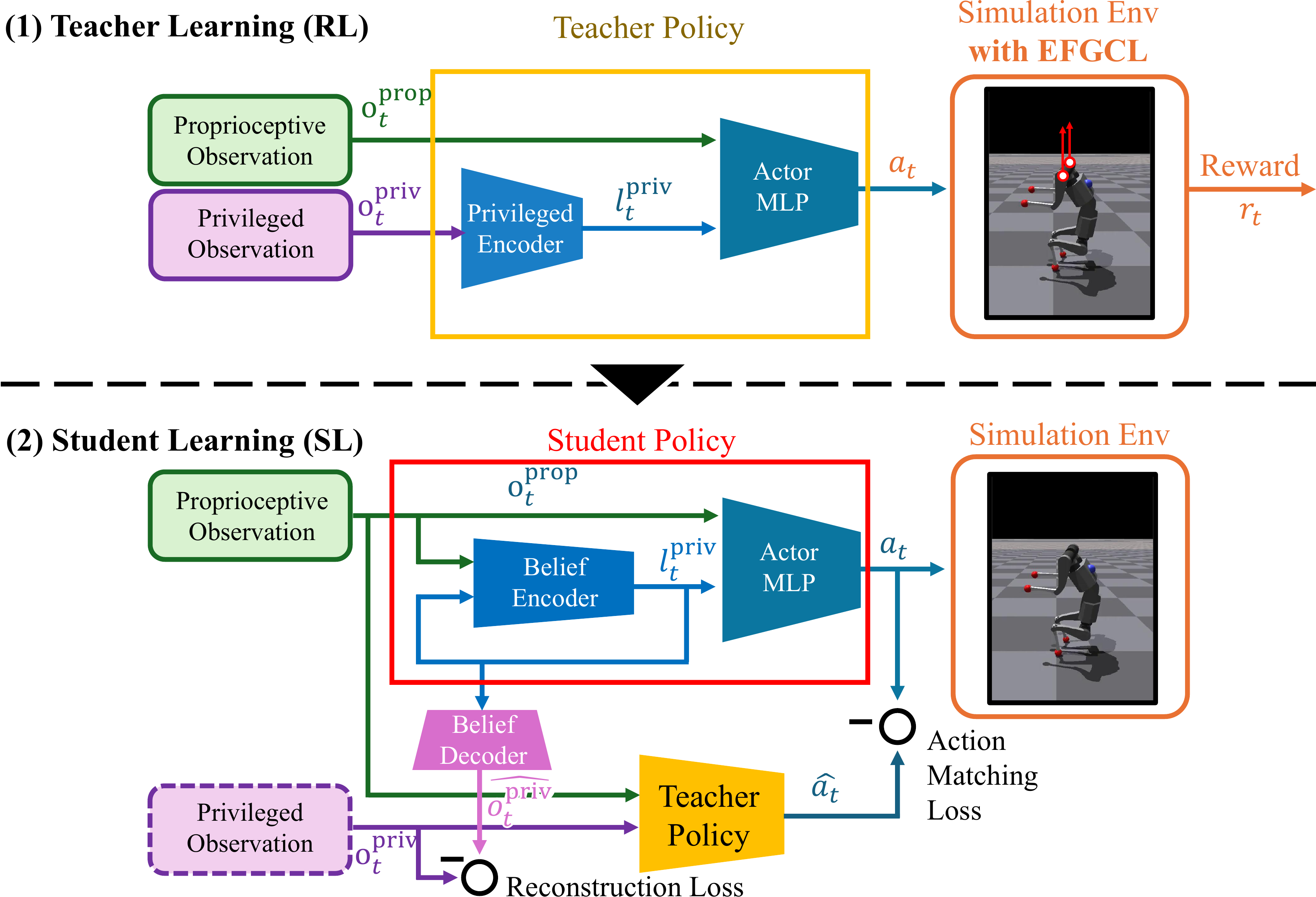}
	\vspace{-1ex}
	\caption{Network architecture used for Teacher--Student learning.
	(a) The Teacher Policy takes the full state, including privileged observations, as input and is trained via reinforcement learning.
	(b) The Student Policy takes only onboard sensor information as input and is trained using the Teacher Policy as a supervisor.}
	\vspace{-3ex}
	\label{fig:network-structure}
\end{figure}

\subsection{Design of External Assistance}\label{sec:design_of_external_assistance}
EFGCL first designs external assistive forces that help reproduce the target motions.
As discussed in \secref{sec:rapid_value_estimation}, exposing the agent to a large number of high-reward trajectories in the early stage of learning induces a tendency to preserve such trajectories.
The external assistance in EFGCL is introduced to artificially increase the density of these successful trajectories.

In EFGCL, external assistance is defined as a pattern consisting of three elements:
the points of application $P=\{\mathbf{p}_i\}$, the corresponding force vectors $F=\{\mathbf{f}_i\}$, and the timing of application
$T=\{(t^{\text{start}}_i, t^{\text{end}}_i)\}$.
Since the purpose of the assistance is not to teach an optimal trajectory but rather to guide the agent toward high-reward states,
it is sufficient for the robot to approximately achieve the target motion and obtain high rewards.
Therefore, in this study, the assistive force $F_{\text{assist}}(P,F,T)$ is heuristically designed for each task.
The permissible range of such assistive forces is investigated in detail in \secref{sec:experiment_ablation}.

\subsection{Success-Rate Based Adaptive Curriculum Scheduling}
As discussed in \secref{sec:curriculum_learning}, curriculum learning benefits from small difficulty gaps between adjacent MDPs to ensure stable policy transitions.
EFGCL adopts a success-rate-based adaptive curriculum to prevent excessive difficulty changes caused by curriculum updates.

Specifically, as shown in lines~6--12 of \algoref{alg:efgcl}, PPO training is repeated at each stage $i$ until the success rate exceeds a threshold $\zeta$.
The assistive force is then updated as
\[
F_i = \alpha_i \times F_{\text{assist}}, \quad
\alpha_i = \max(0, 1 - \varepsilon \times i).
\]
where $\alpha_i$ denotes the assistance scaling factor at stage $i$, 
and $\varepsilon$ is the decay step size controlling the rate of assistance reduction.

This mechanism enables automatic adjustment of the assistance decay step based on training progress, while preserving the relationship
$\pi_{i+1}(a\mid s) / \pi_i(a\mid s) \approx 1$.

\begin{figure}[tbp]
\centering
\includegraphics[width=0.95\columnwidth]{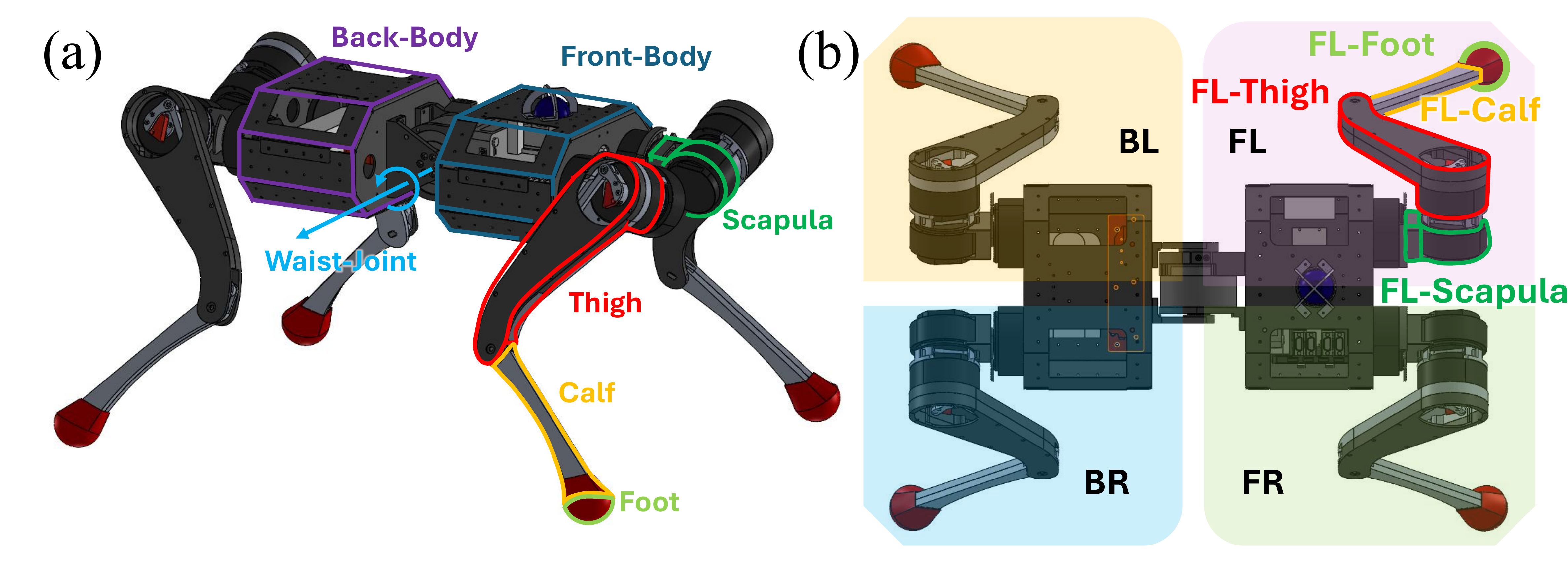}
\vspace{-2ex}
\caption{Overview and kinematic structure of the quadrupedal robot KLEIYN.
(a) Overall robot structure, (b) link and joint definitions for each leg.}
\vspace{-3ex}
\label{fig:robot_model}
\end{figure}

\subsection{Time-Encoding for Observations}
Since the assistive force $F_{\text{assist}}(P,F,T)$ in EFGCL is applied in a time-dependent manner, it is important for the agent to infer the timing of assistance as an internal state.

However, directly using the elapsed time $t$ as an input leads to monotonically increasing values, which may cause scale mismatch in neural networks.

To address this issue, we introduce an additional observation that aligns with the activation interval of the assistive force $F_i$.
Specifically, we define a monotonically increasing function bounded within $[0,1]$ as
\[
\tau(t, \lambda) = \frac{\tilde{t}^3}{1 + \tilde{t}^3}, \quad \tilde{t} = \frac{t}{\lambda}.
\]
Here, $\lambda$ is a temporal scaling parameter, which is set to the force activation start time $\lambda = t^{\text{start}}$ in this study.
}
{
\subsection{Overview of External Force Guided Curriculum Learning (EFGCL)}\label{sec:efgcl_overview}
本研究で提案する External Force Guided Curriculum Learning (EFGCL) は,動的タスクに対して報酬関数を疎に保ちつつ安定した学習を実現する枠組みである.
模倣学習やReward Shapingでは望ましい動作を報酬関数を通じて誘導するが,EFGCLは外力により学習の行われる環境のダイナミクスをカリキュラム的に変化させることで学習の安定化を図る.

アルゴリズム全体の流れを \algoref{alg:efgcl} に示す.
EFGCL は,学習初期にタスク達成を容易にする外部補助力を設計し(line~1),
学習が進むにつれてこの補助を段階的に減衰させる(lines~7--9)カリキュラム学習手法である.
エージェントは,補助力下で成功しやすい軌道を経験しつつ,自律的な動作生成へと徐々に移行し,
最終的には補助なしで目標動作を達成する方策 $\pi_N$ を獲得する(line~14).

\begin{figure}[tbp]
\centering
	\includegraphics[width=0.9\columnwidth]{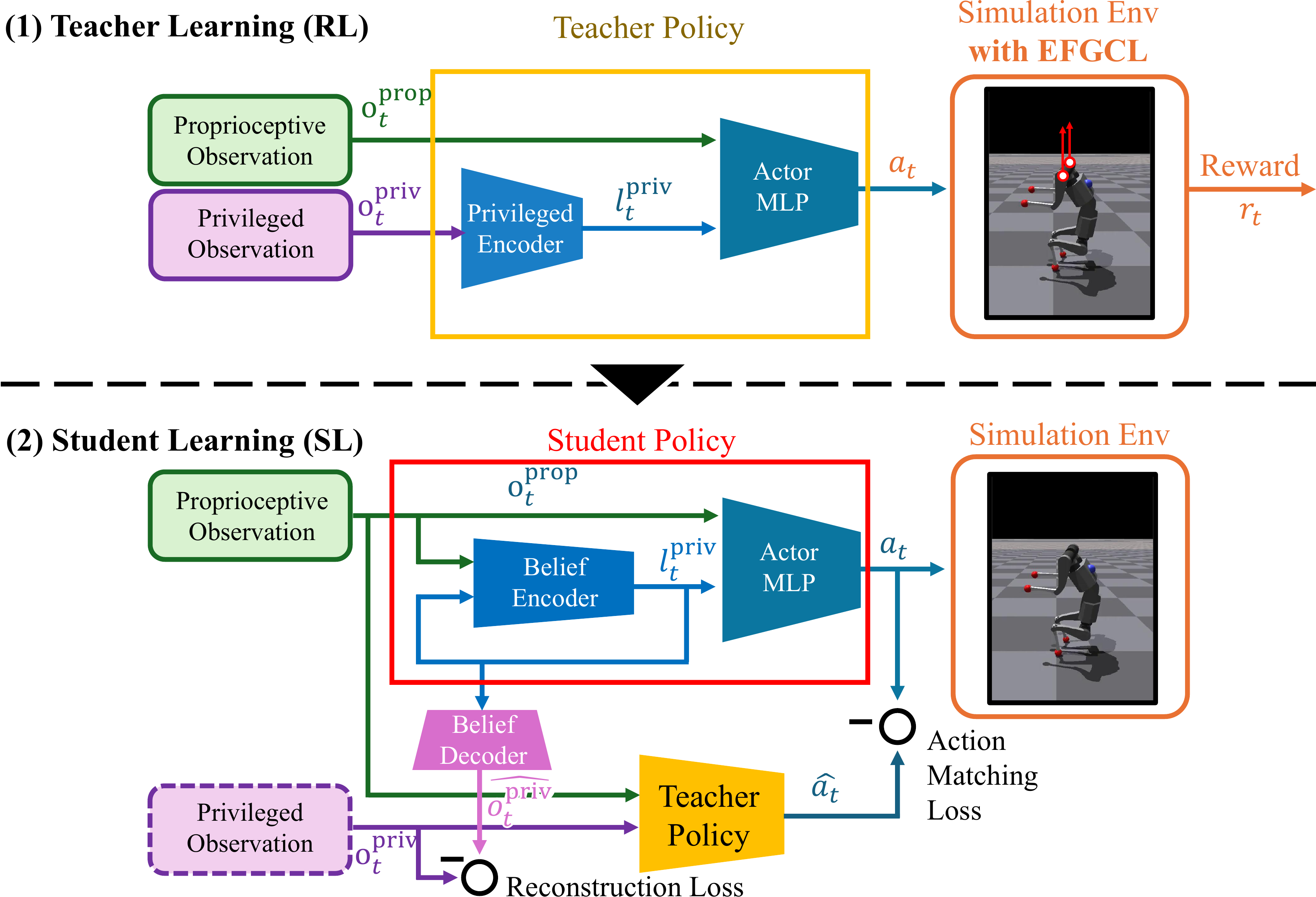}
	\vspace{-1ex}
	\caption{Teacher–Student 学習に用いるネットワーク構成. (a) Teacher Policy は特権的観測を含む完全な状態を入力とし, 強化学習により学習される. (b) Student Policy はオンボードセンサ情報のみを入力とし, Teacher Policy を教師として学習される.}
	\vspace{-3ex}
	\label{fig:network-structure}
\end{figure}

\subsection{Design of External Assistance}\label{sec:design_of_external_assistance}
EFGCL では最初に,目標動作を再現するための外部補助力を設計する.
\secref{sec:rapid_value_estimation}で述べたように,学習初期に報酬が高い軌道を多く経験することによりその軌道を維持しようとする性質が生じる.
EFGCL の外部補助力は,この成功軌道の密度を人工的に高めるために導入される.

EFGCL では外部補助力を作用点$P=\{\mathbf{p}_i\}$とそこに働く力ベクトル$F=\{\mathbf{f}_i\}$ ,そして作用するタイミング$T=\{(t^{\text{start}}_i, t^{\text{end}}_i)\}$ の 3 要素から構成されるパターンと定義する.
これは「最適軌道を教えるため」ではなくあくまで「報酬の高い状態への誘導」として機能するため,目標となる動作をある程度達成し,高い報酬を得られていれば十分である.
そのため本研究ではタスクごとに $F_{\text{assist}}(P,F,T)$ を発見的に設定した
この補助力として許容される範囲については \secref{sec:experiment_ablation} で詳細に検証する.

\subsection{Success-Rate Based Adaptive Curriculum Scheduling}
\secref{sec:curriculum_learning}で述べたように,カリキュラム学習では隣り合う MDP 間の難易度差が小さいほど方策遷移が安定する.
EFGCL では成功率に基づく適応的カリキュラムを採用し,カリキュラムの変化による難易度変化が大きくならないよう調整を行う.

具体的にはアルゴリズム \algoref{alg:efgcl} の 6–12 行目に示すように,各ステージ $i$ において成功率が閾値 $\zeta$ を超えるまで PPO を反復し,その後補助力を
$$F_i = \alpha_i \times F_{\text{assist}}, \quad \alpha_i = \max(0, 1 - \varepsilon \times i)$$
と更新する.

こうすることで, 成功率によって学習が十分に進んだかを測定しつつ補助力減衰のステップを自動調整し,そして$\pi_{i+1}^(a\mid s) / \pi_{i}^(a\mid s) \approx 1$ の関係を維持することが可能となる.

\begin{figure}[tbp]
\centering
\includegraphics[width=0.95\columnwidth]{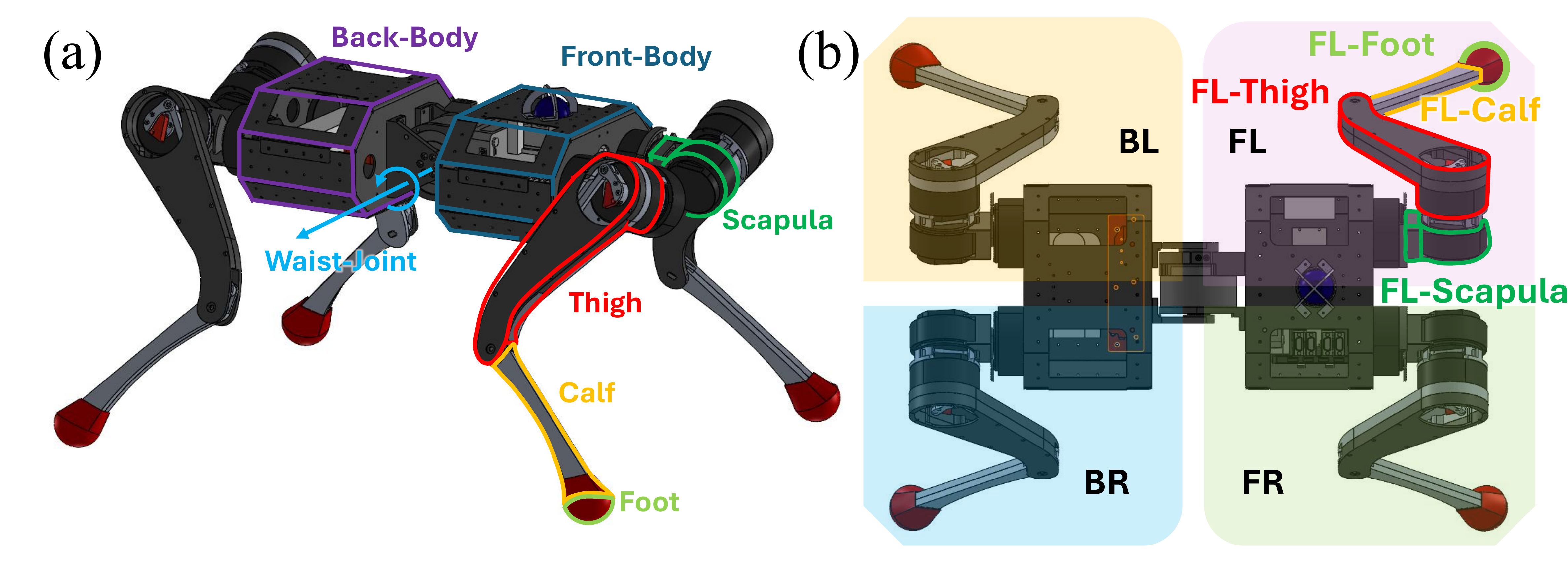}
\vspace{-2ex}
\caption{四脚ロボット KLEIYN の外観とリンク構成. (a) ロボット全体構造, (b) 各脚におけるリンクおよび関節名称.}
\vspace{-3ex}
\label{fig:robot_model}
\end{figure}

\subsection{Time-Encoding for Observations}
EFGCL では補助力 $F_{\text{assist}}(P,F,T)$ が時刻に依存して発生するため,エージェントが補助力のタイミングを内部状態として推測できることが重要となる.

しかし単純な経過時間 $t$ を入力にすると値が単調に増大し,ニューラルネットワークのスケール不整合を引き起こす可能性がある.

そこで本研究では,アルゴリズム中で使用される補助力 $F_i$ の発生区間と整合をとるため,単調増加かつ閾値が[0, 1]に収まる関数として$\tau(t, \lambda) = {\tilde{t}^3}/({1 + \tilde{t}^3}), \quad \tilde{t} = t/\lambda$ を観測として追加した.
ここで $\lambda$ は時間に関するスケールパラメータであり,本研究では力を加え始める時間$\lambda=t^{\text{start}}$ と設定した.

}
{
}

\begin{figure}[htbp]
\centering
	\includegraphics[width=0.9\columnwidth]{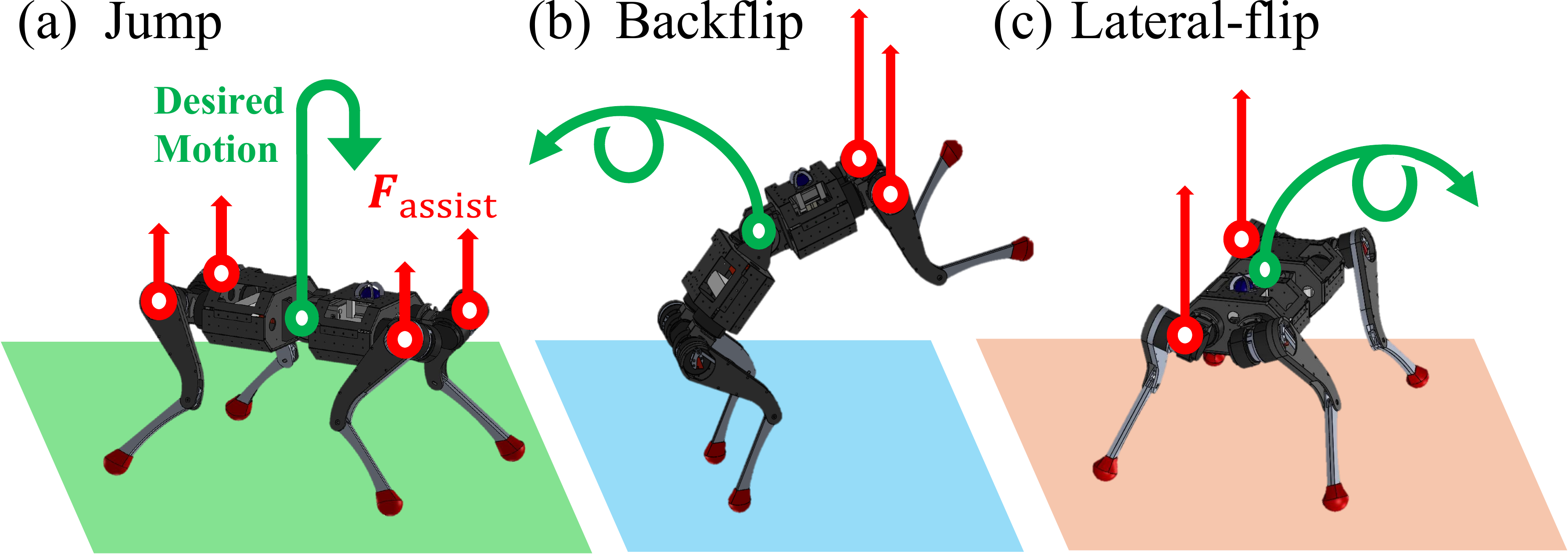}
	\vspace{-1ex}
	\caption{Designed assistive force patterns for each task.
	(a) Jump, (b) Backflip, (c) Lateral-flip.
	The assistive forces are applied vertically to the scapula links, guiding motions that facilitate successful task execution during the early stages of learning.}
	\vspace{-3ex}
	\label{fig:assist-force-pattern}
\end{figure}

\section{Learning Setup}\label{sec:learning_setup}
\switchtext
{

\subsection{Robot Platform}\label{sec:experiment_robot_platform}
The real-world experiments are conducted using the quadrupedal robot KLEIYN~\cite{yoneda2025kleiyn}.
Its appearance and link definitions are shown in \figref{fig:robot_model}.
KLEIYN has a total mass of 18\,kg and a height of 600\,mm, with three degrees of freedom (DoF) per leg and one DoF in the torso.
The robot is equipped with an IMU and joint encoders.
The leg motors are quasi-direct-drive actuators with a maximum torque of 24.8\,Nm, while the torso motor has a maximum torque of 48\,Nm.
Isaac Gym~\cite{liang2018isaacgym} is used as the simulator for training.

\begin{figure*}[tbp]
\centering
	\includegraphics[width=1.9\columnwidth]{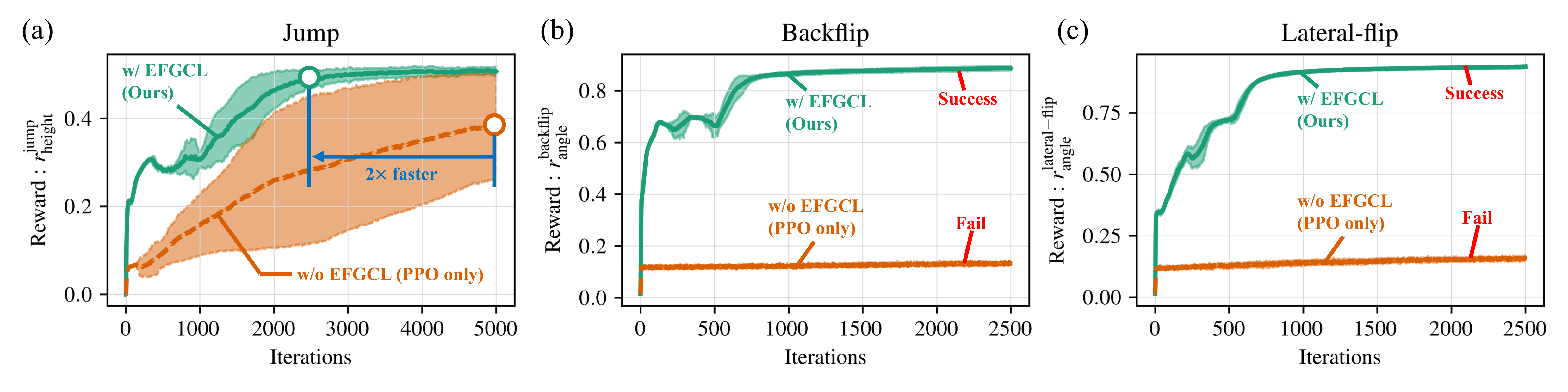}
	\vspace{-1ex}
	\caption{Comparison of learning curves with and without EFGCL.
	(a) Jump, (b) Backflip, (c) Lateral-flip.
	With EFGCL, learning progresses stably and converges to high reward values in all tasks.
	Without EFGCL, Jump exhibits large reward variance, while learning hardly progresses for Backflip and Lateral-flip.}
	\vspace{-3ex}
	\label{fig:reward-comparison}
\end{figure*}

\subsection{Task Definition and Reward Function}
\label{sec:experiment_task_definition}
To evaluate the effectiveness of EFGCL, we define three dynamic whole-body motion tasks:
(1) Jump,
(2) Backflip,
and (3) Lateral-flip.

To avoid arbitrary performance gains due to task-specific reward engineering, all tasks share exactly the same reward structure, weights, and functional forms.
Only the target variables differ between tasks: the maximum height for Jump, and the rotation angle for Backflip and Lateral-flip.
These target variables are simply scaled according to their physical units, and no task-specific reward tuning or intermediate motion-guiding rewards are introduced.

The reward for each task is defined by the following common structure:
\begin{equation}
r_t =
\rho^{\text{task}}_t
+ \rho^{\text{task}}_t \cdot \rho^{\text{stand}}_t
+ \lambda_\omega r^{\text{ang}}_t
+ r^{\text{common}}_t ,
\end{equation}
where $\rho^{\text{task}}_t \in [0,1]$ represents task progress,
$\rho^{\text{stand}}_t$ encourages stable posture after landing,
$r^{\text{ang}}_t$ is a regularization term that suppresses rotation about non-target axes,
and $r^{\text{common}}_t$ enforces physical constraints shared across all tasks.

The only task-specific difference lies in the definition of the target quantity in $\rho^{\text{task}}_t$.
For Jump, the target is the maximum achieved height, while for Backflip and Lateral-flip, the targets are the rotation angles around the pitch and roll axes, respectively.
Detailed definitions are summarized in Appendix~\ref{app:reward_detail}.

\subsection{Observations}\label{sec:observations}
The observation consists of two components, $\mathbf{o}^{\text{prop}}_t$ and $\mathbf{o}^{\text{priv}}_t$.

$\mathbf{o}^{\text{prop}}_t$ represents proprioceptive observations and includes joint positions $\mathbf{q}_t\in\mathbb{R}^{13}$,
joint velocities $\dot{\mathbf{q}}_t\in\mathbb{R}^{13}$,
the gravity vector in the root frame $\tilde{\mathbf{g}}_t\in\mathbb{R}^3$,
root angular velocity $\boldsymbol{\omega}_t\in\mathbb{R}^3$,
the command input $c\in\mathbb{R}$,
and the time encoding $\tau(t)\in\mathbb{R}$.
The command input corresponds to the target jump height $h^{\text{target}}$ for Jump,
and the target rotation angle $\theta^{\text{target}}$ for Backflip and Lateral-flip.

$\mathbf{o}^{\text{priv}}_t$ denotes privileged observations used exclusively by the Teacher Policy and consists of the following task-specific information:
\begin{itemize}
	\item Jump: root height $h_t\in\mathbb{R}$ and maximum height since episode start $h_t^{\text{max}}\in\mathbb{R}$
	\item Backflip: root height $h_t$ and root pitch angle $\theta^{\text{pitch}}_t\in\mathbb{R}$
	\item Lateral-flip: root height $h_t$ and root roll angle $\theta^{\text{roll}}_t\in\mathbb{R}$
\end{itemize}

\subsection{Assist Force Design}\label{sec:experiment_assist_force_design}
Assistive force patterns are designed for each task.
The application timing is shared across all tasks as
$T = \{(1.0\,\mathrm{s}, 1.1\,\mathrm{s})\}$,
while the application points $P$ and force vectors $F$ are task-specific.
The assistive force patterns are illustrated in \figref{fig:assist-force-pattern}.

\paragraph{Jump}
\[
\begin{aligned}
	P^{\text{jump}} &= \{\mathbf{p}^{\text{Scapula}}_{\text{FL}}, \mathbf{p}^{\text{Scapula}}_{\text{FR}}, \mathbf{p}^{\text{Scapula}}_{\text{BL}}, \mathbf{p}^{\text{Scapula}}_{\text{BR}}\},\\
	F^{\text{jump}} &= \{(0, 0, f_{\text{jump}}(h^{\text{target}})/4)\}
\end{aligned}
\]
Here, $f_{\text{jump}}(h^{\text{target}})$ is the assistive force magnitude determined by the target height $h^{\text{target}}$.
This value is derived by modeling the Jump motion as simple projectile motion and computing the average assistive force required to generate the initial velocity needed to reach the target height.

\paragraph{Backflip}
\[
\begin{aligned}
	P^{\text{backflip}} &= \{\mathbf{p}^{\text{Scapula}}_{\text{FL}}, \mathbf{p}^{\text{Scapula}}_{\text{FR}}\},\\
	F^{\text{backflip}} &= \{(0, 0, 175\,\mathrm{N})\}
\end{aligned}
\]

\paragraph{Lateral-flip}
\[
\begin{aligned}
	P^{\text{lateral}} &= \{\mathbf{p}^{\text{Scapula}}_{\text{FR}}, \mathbf{p}^{\text{Scapula}}_{\text{BR}}\},\\
	F^{\text{lateral}} &= \{(0, 0, 300\,\mathrm{N})\}
\end{aligned}
\]

\subsection{Adaptive Curriculum Design}\label{sec:experiment_curriculum_design}
A success-rate-based curriculum scheduling strategy is employed to decay the assistive forces.
The success rate in \algoref{alg:efgcl} is computed using the following criteria.

\paragraph{Jump}
\[
|h_t^{\text{max}} - h^{\text{target}}| < 0.1 \ \land \ |h_t| < 0.1
\]

\paragraph{Backflip, Lateral-flip}
\[
|\theta_t - 2\pi| < 0.3 \ \land \ |h_t| < 0.1
\]

The success rate threshold is set to $\zeta = 0.6$, and the decay step size is set to $\varepsilon = 0.01$ for all tasks.

\subsection{Teacher--Student Learning}
Markovianity is a crucial property in reinforcement learning environments.
Following prior work~\cite{miki2022anymal}, we adopt a Teacher--Student learning architecture.

Teacher--Student learning consists of two stages: reinforcement learning of the teacher policy and supervised learning of the student policy, where the teacher policy acts as a supervisor.
During supervised learning, the student policy is trained to minimize an action-matching loss and a reconstruction loss on privileged observations.
Since the proposed method is integrated with reinforcement learning, EFGCL is applied only during the training of the teacher policy.
The overall learning structure is illustrated in \figref{fig:network-structure}.
}
{
本章では,提案手法 EFGCL(External Force Guided Curriculum Learning)の有効性を検証するために,シミュレーションおよび実機実験による評価を行う.
本研究の実験は以下の 3 つを目的として構成される:

\begin{itemize}
  \item[(1)] EFGCL が学習の安定化と高速化に寄与するか
  \item[(2)] 学習した方策が実機四足ロボットで再現可能か
  \item[(3)] 補助力パターンの設計が学習性能に与える影響
\end{itemize}

\subsection{Robot Platform}\label{sec:experiment_robot_platform}
実機には四脚ロボットKLEIYN~\cite{yoneda2025kleiyn}を用いた.
外観と各リンクの名称を\figref{fig:robot_model}に示す.
KLEIYNは質量18\,kg,全高600\,mm,脚部に3-DoF, 腹部に1-DoFを有する四脚ロボットである.
IMUと関節エンコーダを搭載しており,脚部のモータは最大24.8\, Nm, 胴体のモータは最大48\, NmのQuasi-Direct-Drive モータである.
また学習を行うシミュレータにはIsaac Gym\cite{liang2018isaacgym}を用いた.
\begin{figure}[bp]
\centering
	\includegraphics[width=0.9\columnwidth]{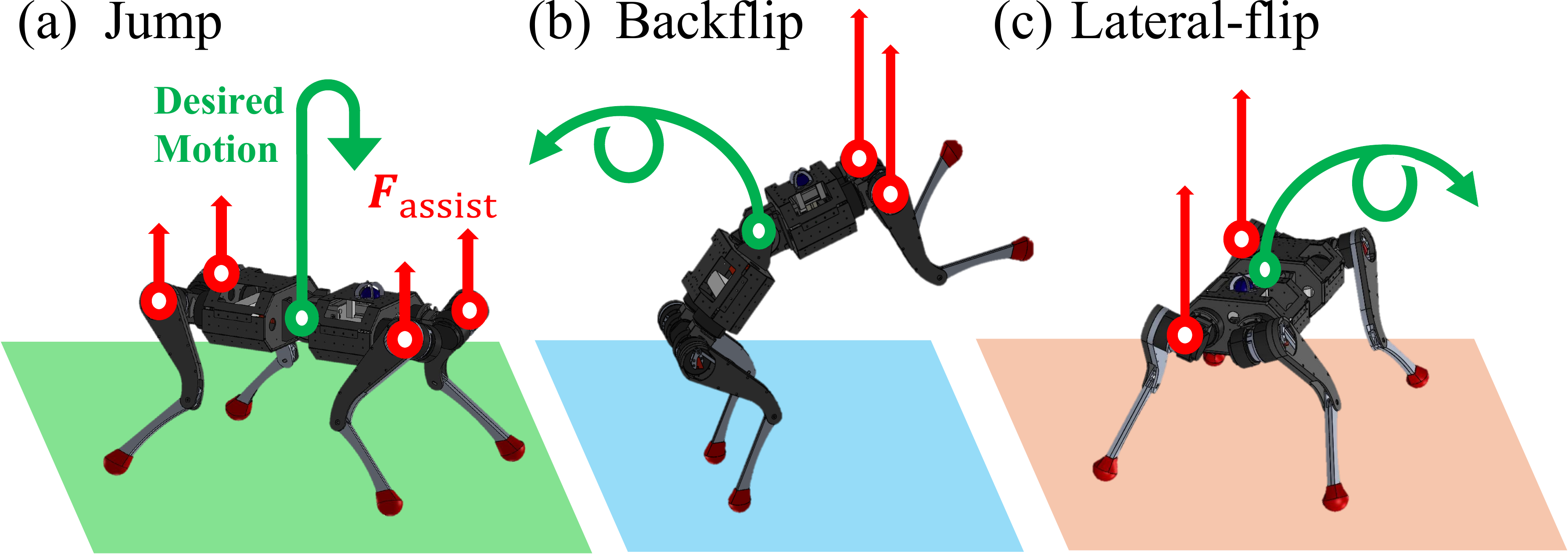}
	\vspace{-1ex}
	\caption{ 各タスクにおいて設計した補助力パターン. (a) Jump, (b) Backflip, (c), Lateral-flip. 補助力はScapula リンクに対して鉛直方向に作用し, 学習初期に目標動作が成立しやすい運動を誘導する.}
	\vspace{-3ex}
	\label{fig:assist-force-pattern}
\end{figure}

\begin{figure*}[tbp]
\centering
	\includegraphics[width=1.9\columnwidth]{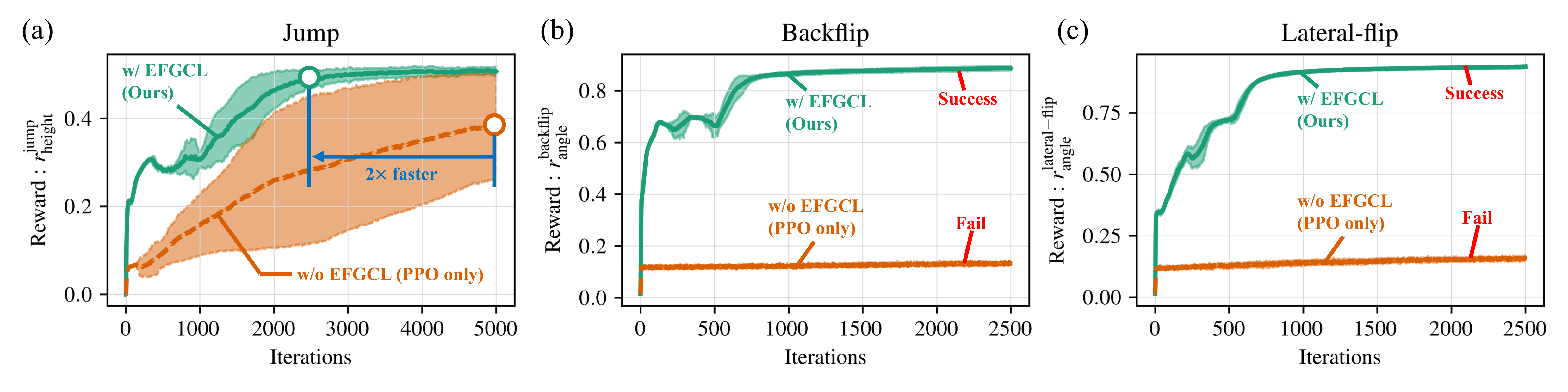}
	\vspace{-1ex}
	\caption{EFGCL の有無による学習曲線の比較. (a) Jump,(b) Backflip,(c) Lateral-Flip. EFGCL を適用した場合,すべてのタスクで学習が安定して進行し, 高い報酬値に収束する.一方,EFGCL を用いない場合, Jump では報酬の分散が大きく,Backflip および Lateral-Flip では 学習がほとんど進行しない.}
	\vspace{-3ex}
	\label{fig:reward-comparison}
\end{figure*}

\subsection{Task Definition and Reward Function}
\label{sec:experiment_task_definition}
EFGCLの有効性を検証するために, 以下の3種類のダイナミックな全身運動タスクを設定した:
(1) 垂直跳び(Jump), 
(2) バク転(Backflip), 
(3) 側転(Lateral-Flip).

本研究では, タスク固有の報酬設計による恣意的な性能向上を避けるため, すべてのタスクにおいて報酬関数の構造, 重み, および関数形を完全に共通とした.
タスク間で変更されるのは目標変数のみであり, Jumpでは到達高さ, BackflipおよびLateral-Flipでは回転角が対応する.
これらは物理量に応じて単純にスケーリングされるのみであり, タスクごとの報酬チューニングや, 目標動作に至る中間動作を明示的に誘導する報酬は一切導入していない.

各タスクにおける報酬は, 以下の共通構造を持つ:
\begin{equation}
r_t =
\rho^{\text{task}}_t
+ \rho^{\text{task}}_t \cdot \rho^{\text{stand}}_t
+ \lambda_\omega r^{\text{ang}}_t
+ r^{\text{common}}_t ,
\end{equation}
ここで $\rho^{\text{task}}_t$ は0から1でタスク進行度を表す報酬項,
$\rho^{\text{stand}}_t$ は着地後の姿勢安定性を促す報酬項,
$r^{\text{ang}}_t$ は非目標軸回転を抑制する正則化項である.
$r^{\text{common}}_t$ は全タスク共通の物理的制約に関する報酬である.

タスクごとの違いは $\rho^{\text{task}}_t$ における目標量の定義のみであり,
Jumpでは最大到達高さ, BackflipおよびLateral-Flipではそれぞれピッチ軸・ロール軸周りの回転角を目標とする.
これらの具体的な定義は にてTable~\ref{tab:task_reward} にまとめる.

\subsection{Observations}\label{sec:observations}
観測情報は $\mathbf{o}^{\text{prop}}_t, \mathbf{o}^{\text{priv}}_t$ の2種類から構成される.

$\mathbf{o}^{\text{prop}}_t$ は四脚ロボットの自己観測情報であり, 関節角度$\mathbf{q}_t\in\mathbb{R}^{13}$, 関節速度$\mathbf{\dot{q}}_t\in\mathbb{R}^{13}$, ルートリンクの重力ベクトル$\mathbf{\tilde{g}}_t\in\mathbb{R}^3$, ルートリンクの角速度$\omega_t\in\mathbb{R}^3$, コマンド入力$c\in\mathbb{R}$, time-encoding $\tau(t)\in\mathbb{R}$から構成される. 
コマンド入力はジャンプタスクでは目標ジャンプ高さ$h^{\text{target}}\in\mathbb{R}$であり, バク転および側転タスクでは目標回転角度$\theta^{\text{target}}\in\mathbb{R}$である.

$\mathbf{o}^{\text{priv}}_t$ はTeacher Policy専用の観測情報であり,タスクごとに以下の情報から構成される.
\begin{itemize}
	\item Jump: ルートリンクの高さ$h_t\in\mathbb{R}$, エピソード開始からの最高到達高さ$h_t^{\text{max}}\in\mathbb{R}$
	\item Backflip:  ルートリンクの高さ$h_t$, ルートリンクのピッチ角$\theta^{\text{pitch}}_t\in\mathbb{R}$
	\item Lateral-Flip: ルートリンクの高さ$h_t$, ルートリンクのロール角$\theta^{\text{roll}}_t\in\mathbb{R}$
\end{itemize}

\subsection{Assist Force Design}\label{sec:experiment_assist_force_design}
各タスクに対して補助力パターンを設計した.
全タスクで作用タイミングは共通で $T = \{(1.0 \mathrm{s}, 1.1 \mathrm{s})\}$ とし,タスクごとに以下の作用点$P$力ベクトル$F$を設計した.
各タスクでの補助力パターンを\figref{fig:assist-force-pattern}に示す.
\paragraph{Jump}
$$
\begin{align*}
	P^{\text{jump}} &= \{\mathbf{p}^{\text{Scapula}}_{\text{FL}}, \mathbf{p}^{\text{Scapula}}_{\text{FR}}, \mathbf{p}^{\text{Scapula}}_{\text{BL}}, \mathbf{p}^{\text{Scapula}}_{\text{BR}}\}\\
	F^{\text{jump}} &= \{(0, 0, f_{\text{jump}}(h^{\text{target}})/4)\}
\end{align*}
$$
ここで$f_{\text{jump}}(h^{\text{target}})$は目標高さ$h^{\text{target}}$に応じて設定される補助力である.
この導出はジャンプ動作を単純な放物運動としてモデル化し,目標高さに到達するために必要な初速度を得るための平均的な補助力を計算することで得られる.

\paragraph{Backflip}
$$
\begin{align*}
	P^{\text{backflip}} &= \{\mathbf{p}^{\text{Scapula}}_{\text{FL}}, \mathbf{p}^{\text{Scapula}}_{\text{FR}}\}\\
	F^{\text{backflip}} &= \{(0, 0, 175 \ \mathrm{N}))\}
\end{align*}
$$

\paragraph{Lateral-Flip}
$$
\begin{align*}
	P^{\text{backflip}} &= \{\mathbf{p}^{\text{Scapula}}_{\text{FR}}, \mathbf{p}^{\text{Scapula}}_{\text{BR}}\}\\
	F^{\text{backflip}} &= \{(0, 0, 300 \ \mathrm{N}))\}
\end{align*}
$$

\subsection{Adaptive Curriculum Design}\label{sec:experiment_curriculum_design}
補助力の減衰スケジューリングには,成功率を用いたカリキュラムスケジューリングを用いた.
\algoref{alg:efgcl}におけるSuccess\_Rateの計算には,以下の成功条件を用いた.
\paragraph{Jump}
$$|h_t^{\text{max}}-h^{\text{target}}| < 0.1 \ \& \ |h_t| < 0.1$$
\paragraph{Backflip, Lateral-Flip}
$$|\theta_t-2\pi| < 0.3 \ \& \ |h_t| < 0.1$$
また各タスクでの成功率閾値$\zeta=0.6$, 減衰ステップ幅$\varepsilon=0.01$とした.

\subsection{Teacher-Student Learning Architecture}
一般に強化学習を行う環境ではマルコフ性が大事である.
そこで本研究では,既存研究 \cite{miki2022anymal} に基づき Teacher-Student 構造を採用した.
Teacher-Student学習は, Teacher-Policyの強化学習とTeacher-Policyを教師としたStudent-Policyの教師有り学習の2段階による学習手法である.

本手法は強化学習と組み合わせる手法であるため, Teacher-Policyの学習のみに適用した.
学習の構造を \figref{fig:network-structure} に示す.
}
{
}

\section{Experiment and Result}\label{sec:experiment_and_result}
\switchtext
{
\subsection{Learning Performance and Real Robot Deployment}\label{sec:experiment_learning_compare}
We compared the proposed EFGCL with a PPO baseline over 10 random seeds.
As shown in \figref{fig:reward-comparison}, EFGCL achieved stable convergence and high rewards across all tasks (Jump, Backflip, and Lateral-flip).
In contrast, the baseline failed to learn the flipping tasks and exhibited high variance in the Jump task.
As visualized in \figref{fig:result-sim}, the baseline often resulted in unnatural postures, whereas EFGCL acquired natural dynamic motions.
This stable learning process was facilitated by the adaptive curriculum, which automatically adjusted the assistive force decay based on the success rate (\figref{fig:efgcl-value-transition}).
Furthermore, the policies learned via EFGCL were distilled and deployed on the quadrupedal robot KLEIYN.
As shown in \figref{fig:jump_real}, the dynamic motions observed in simulation were successfully reproduced on the real robot for all three tasks.

\begin{figure}[tbp]
\centering
	\includegraphics[width=0.9\columnwidth]{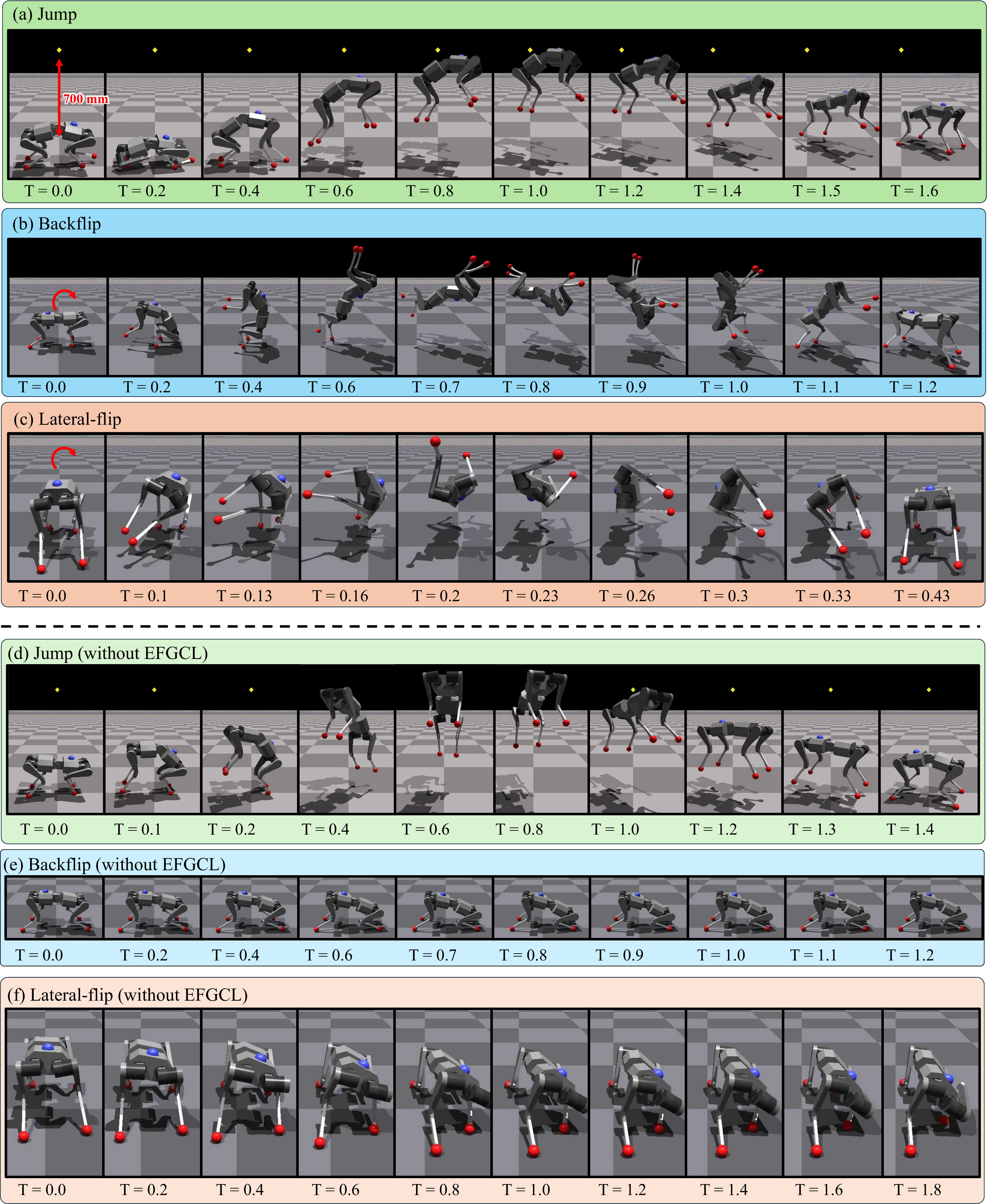}
	\vspace{-1ex}
	\caption{Snapshots of learned motions with and without EFGCL.
	(a--c) With EFGCL, (d--f) without EFGCL.
	With EFGCL, natural and stable motions are acquired, whereas without EFGCL, unnatural postures and failure to achieve the target motions are observed.}
	\vspace{-1ex}
	\label{fig:result-sim}
\end{figure}

\begin{figure}[tbp]
\centering
	\includegraphics[width=0.9\columnwidth]{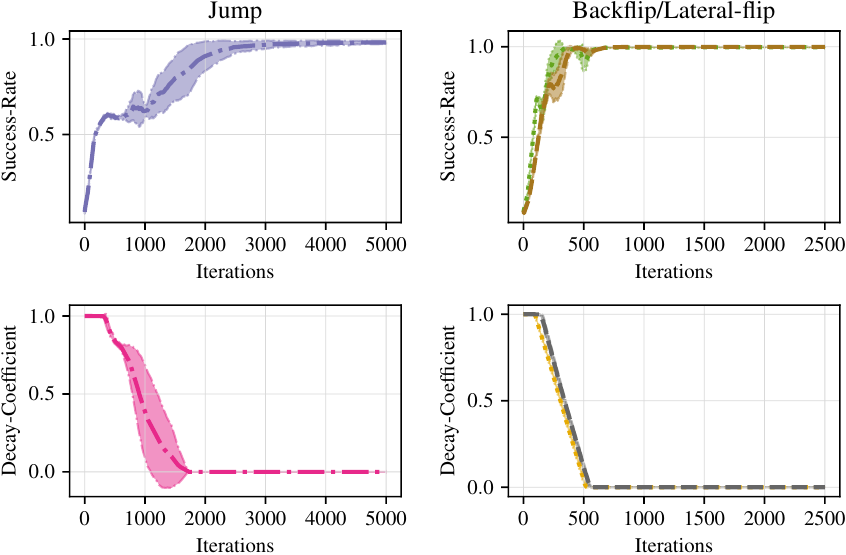}
	\vspace{-1ex}
	\caption{Transitions of the success rate and assistive force decay factor in EFGCL.
	(a) Jump task, (b) Backflip and Lateral-flip tasks.
	The decay speed of the assistive force is automatically adjusted according to the success rate.}
	\vspace{-1ex}
	\label{fig:efgcl-value-transition}
\end{figure}

\begin{figure}[tbp]
\centering
	\includegraphics[width=0.9\columnwidth]{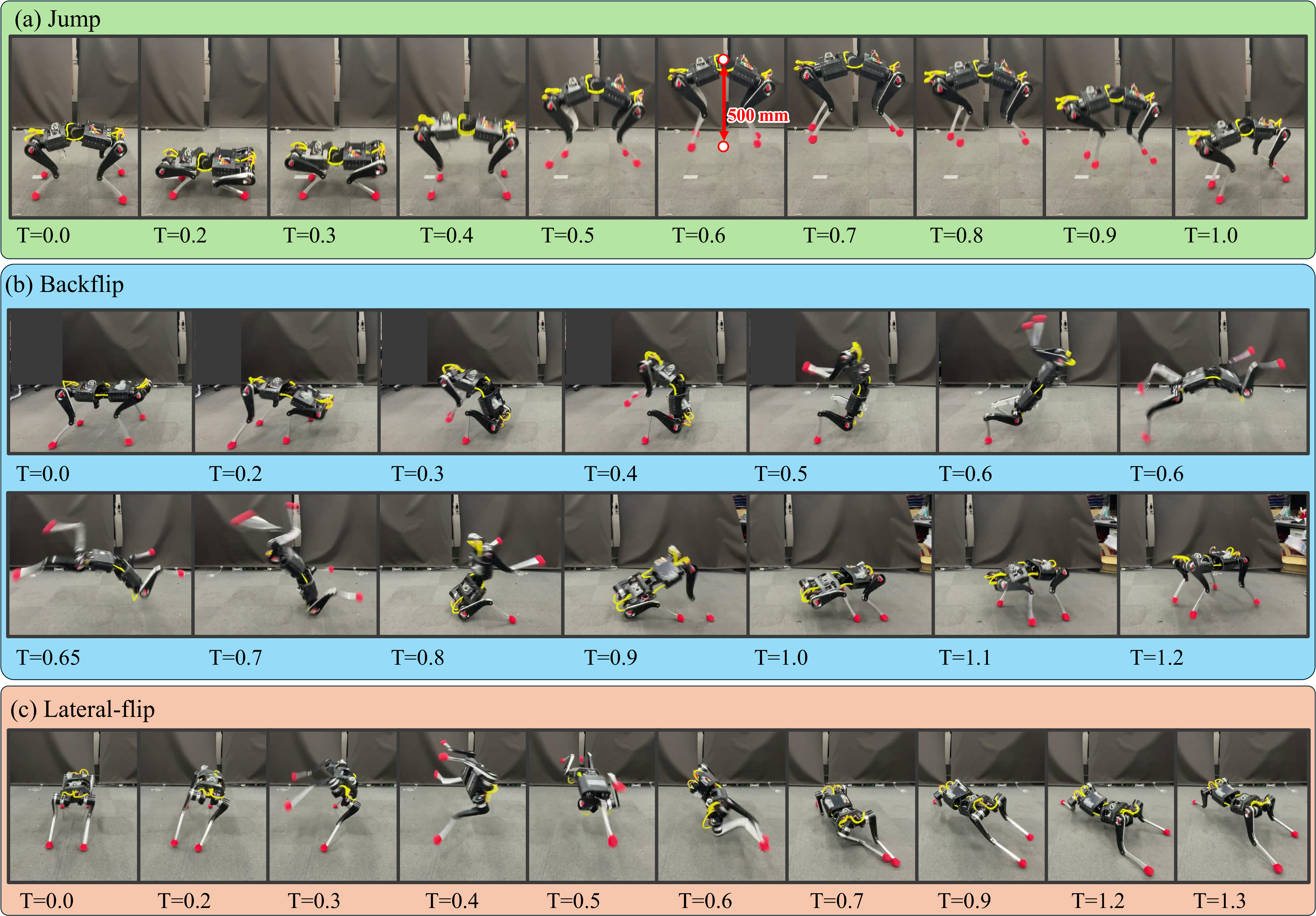}
	\vspace{-1ex}
	\caption{Reproduction of learned motions on the real quadrupedal robot.
	(a) Jump, (b) Backflip, (c) Lateral-flip.}
	\vspace{-1ex}
	\label{fig:jump_real}
\end{figure}

\begin{figure}[tbp]
\centering
	\includegraphics[width=0.9\columnwidth]{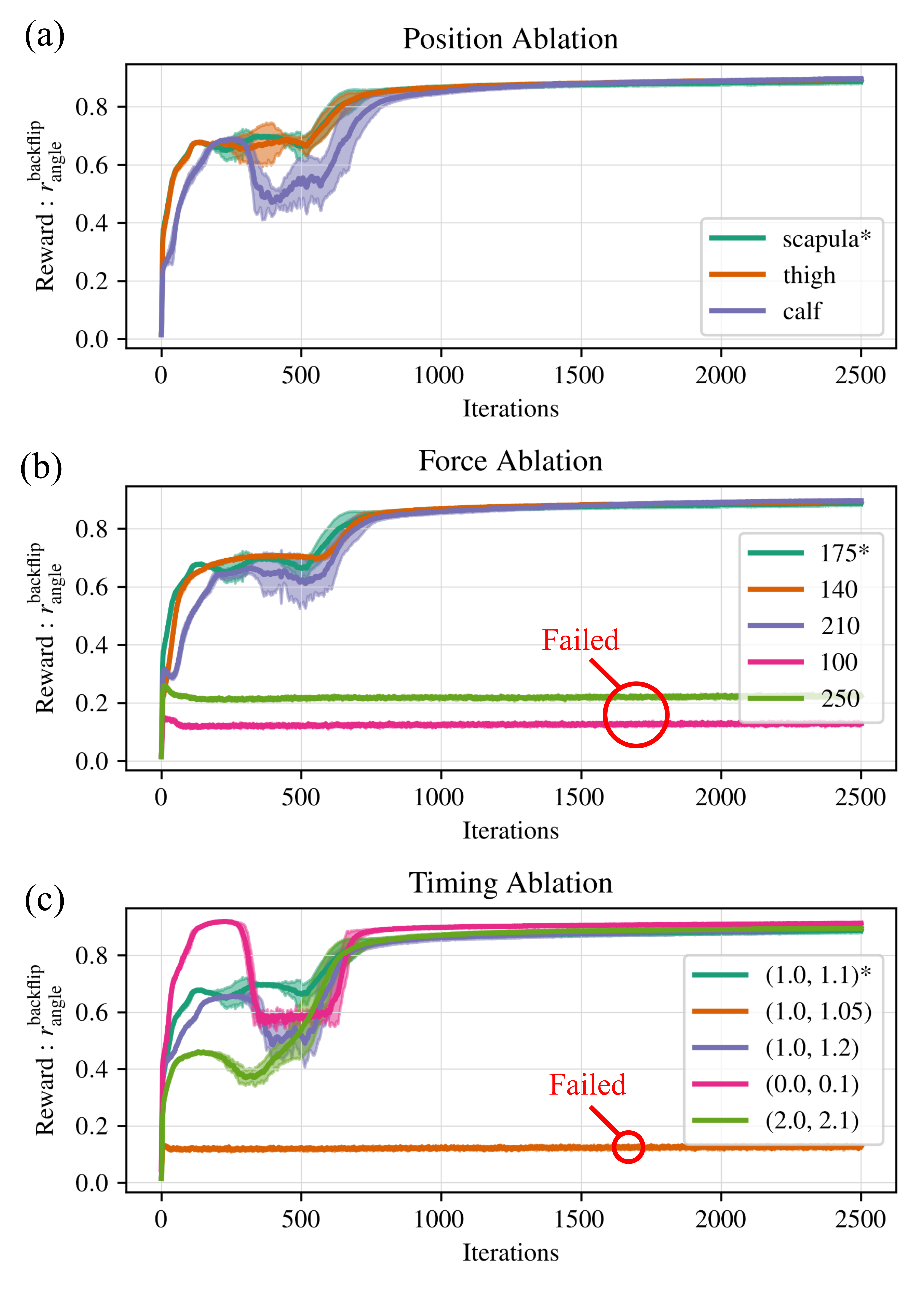}
	\caption{Ablation study on assistive force design.
	(a) Application points, (b) force magnitudes, (c) application timing.
	The asterisks (*) in the legends indicate the conditions used in the main experiments.
	}
	\vspace{-3ex}
	\label{fig:assist-force-ablation}
\end{figure}

\subsection{Ablation Study of EFGCL Force Design}\label{sec:experiment_ablation}
To evaluate the sensitivity of EFGCL to heuristic design choices, we varied the application point, magnitude, and timing of the assistive force in the Backflip task.
The results in \figref{fig:assist-force-ablation} demonstrate that learning is robust over a wide range of parameters.
Successful policies were acquired even when the force was applied to different links (thigh or calf) or when the magnitude varied within a reasonable range (140--210\,N).
Learning failed only in extreme cases where the assistance was physically insufficient (e.g., 100\,N) or excessive (e.g., 250\,N), or when the application timing was too short (1.0\,s to 1.05\,s).
These results indicate that precise tuning is not required, as long as the assistance roughly facilitates the target motion.

\subsection{Evaluation of Accelerated Critic Value Estimation}\label{sec:experiment_value_function_compare}
To validate the hypothesis that external guidance accelerates critic learning, we analyzed value estimates during the Jump task.
\figref{fig:value_function_compare} compares the value function outputs for a successful reference motion at different training stages.
With EFGCL, the value estimates converged to the final distribution as early as 200 iterations.
In contrast, the baseline required more than 1{,}000 iterations to reach a comparable level of accuracy and exhibited larger variance.
This result confirms that experiencing successful states early in training significantly accelerates value function estimation.

\begin{figure}[tbp]
\centering
	\vspace{-5ex}
	\includegraphics[width=0.95\columnwidth]{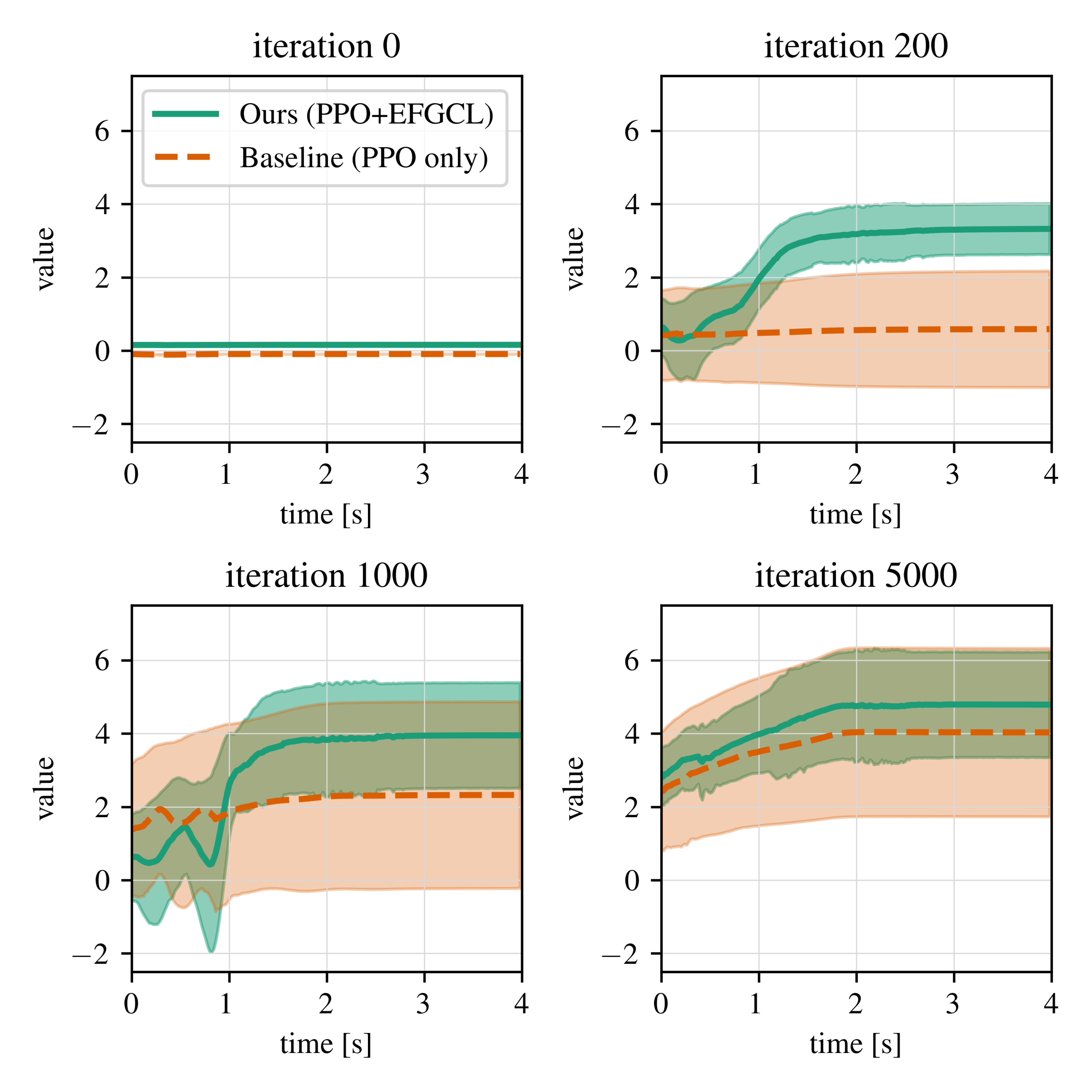}
	\vspace{-3ex}
	\caption{Comparison of value function estimation with and without EFGCL.
	With EFGCL, value estimates quickly converge to distributions close to the final value function, demonstrating accelerated and stabilized value estimation.}
	\vspace{-3ex}
	\label{fig:value_function_compare}
\end{figure}

}
{
\subsection{学習性能および実機ロボットへの適用}\label{sec:experiment_learning_compare}

提案手法 EFGCL と PPO ベースラインを,10 個のランダムシードにおいて比較した.
\figref{fig:reward-comparison} に示すように,EFGCL は Jump,Backflip,Lateral-flip のすべてのタスクにおいて安定した収束と高い報酬値を達成した.
一方で,ベースラインはフリップ系タスクの学習に失敗し,Jump タスクにおいても学習の分散が大きかった.

\figref{fig:result-sim} に示すように,ベースラインでは不自然な姿勢に陥る例が多く観察されたのに対し,EFGCL では自然で動的な運動が獲得された.
この安定した学習過程は,成功率に応じて補助力の減衰を自動調整する適応的カリキュラムによって支えられている(\figref{fig:efgcl-value-transition}).

さらに,EFGCL により学習された方策を蒸留し,四脚ロボット KLEIYN に実装した.
\figref{fig:jump_real} に示すように,シミュレーション上で獲得された動的運動は,すべてのタスクにおいて実機ロボット上でも安定して再現された.\begin{figure}[tbp]

\centering
	\includegraphics[width=0.9\columnwidth]{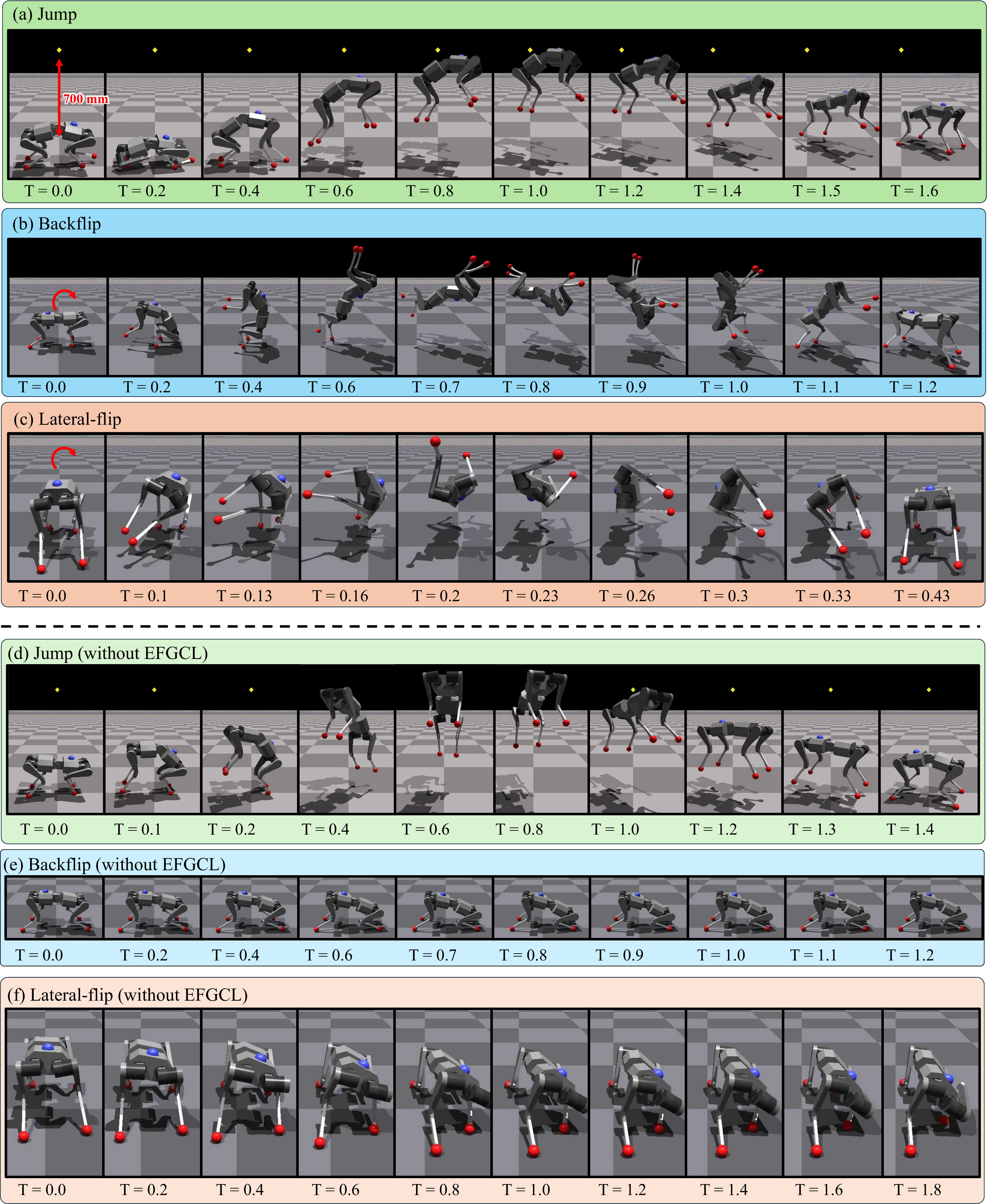}
	\vspace{-1ex}
	\caption{EFGCL の有無による学習動作のスナップショット. (a,b,c) EFGCL あり,(d–f) EFGCL なし. EFGCL を用いた場合は自然で安定した動作が獲得されているのに対し, EFGCL を用いない場合は不自然な姿勢や目標動作の未達が見られる.}
	\vspace{-1ex}
	\label{fig:result-sim}
\end{figure}
\begin{figure}[tbp]
\centering
	\includegraphics[width=0.9\columnwidth]{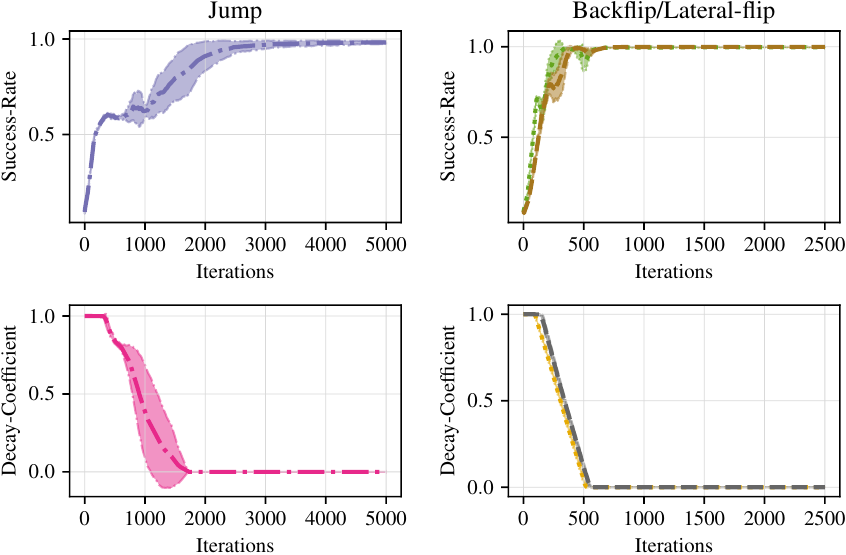}
	\vspace{-1ex}
	\caption{EFGCL における成功率および補助力減衰係数の推移. (a) Jump タスク,(b) Backflip／Lateral-Flip タスク. 成功率に応じて補助力の減衰速度が自動的に調整されていることが分かる.}
	\vspace{-1ex}
	\label{fig:efgcl-value-transition}
\end{figure}

\figref{fig:assist-force-ablation}に各条件での学習結果を示す.
\begin{figure}[tbp]
\centering
	\includegraphics[width=0.9\columnwidth]{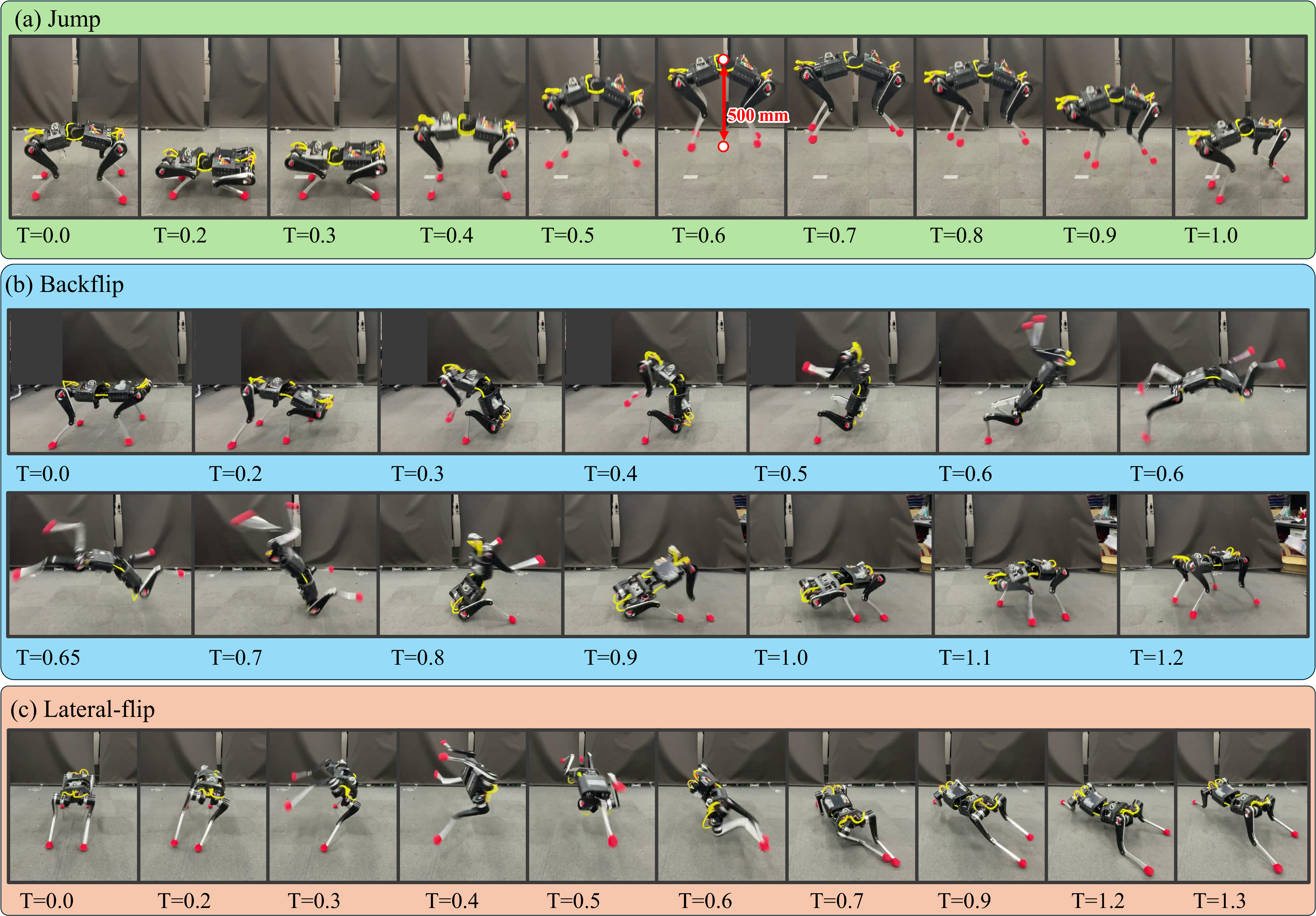}
	\vspace{-1ex}
	\caption{実機四脚ロボットにおける学習済み動作の再現結果. (a) Jump,(b) Backflip,(c) Lateral-Flip. シミュレーションで学習した動作が,すべてのタスクにおいて 実機上でも安定して再現可能であることを示す.}
	\vspace{-3ex}
	\label{fig:jump_real}
\end{figure}

\subsection{EFGCL における外力設計のアブレーションスタディ}\label{sec:experiment_ablation}

EFGCL におけるヒューリスティックな設計要素への感度を評価するため,Backflip タスクにおいて補助力の作用点,大きさ,および付与タイミングを変更した.
\figref{fig:assist-force-ablation} に示す結果から,学習は広いパラメータ範囲において安定して成立することが確認された.

具体的には,作用点を大腿部や下腿部に変更した場合や,補助力の大きさを 140–210,N の範囲で変化させた場合でも,成功する方策が獲得された.
一方で,補助力が物理的に不十分な場合(100,N)や過剰な場合(250,N),あるいは付与時間が極端に短い場合(1.0–1.05,s)には学習が失敗した.
これらの結果は,補助力が目標動作を大まかに支援するものであれば,厳密なチューニングを必要としないことを示している.

\begin{figure}[tbp]
\centering
	\includegraphics[width=0.9\columnwidth]{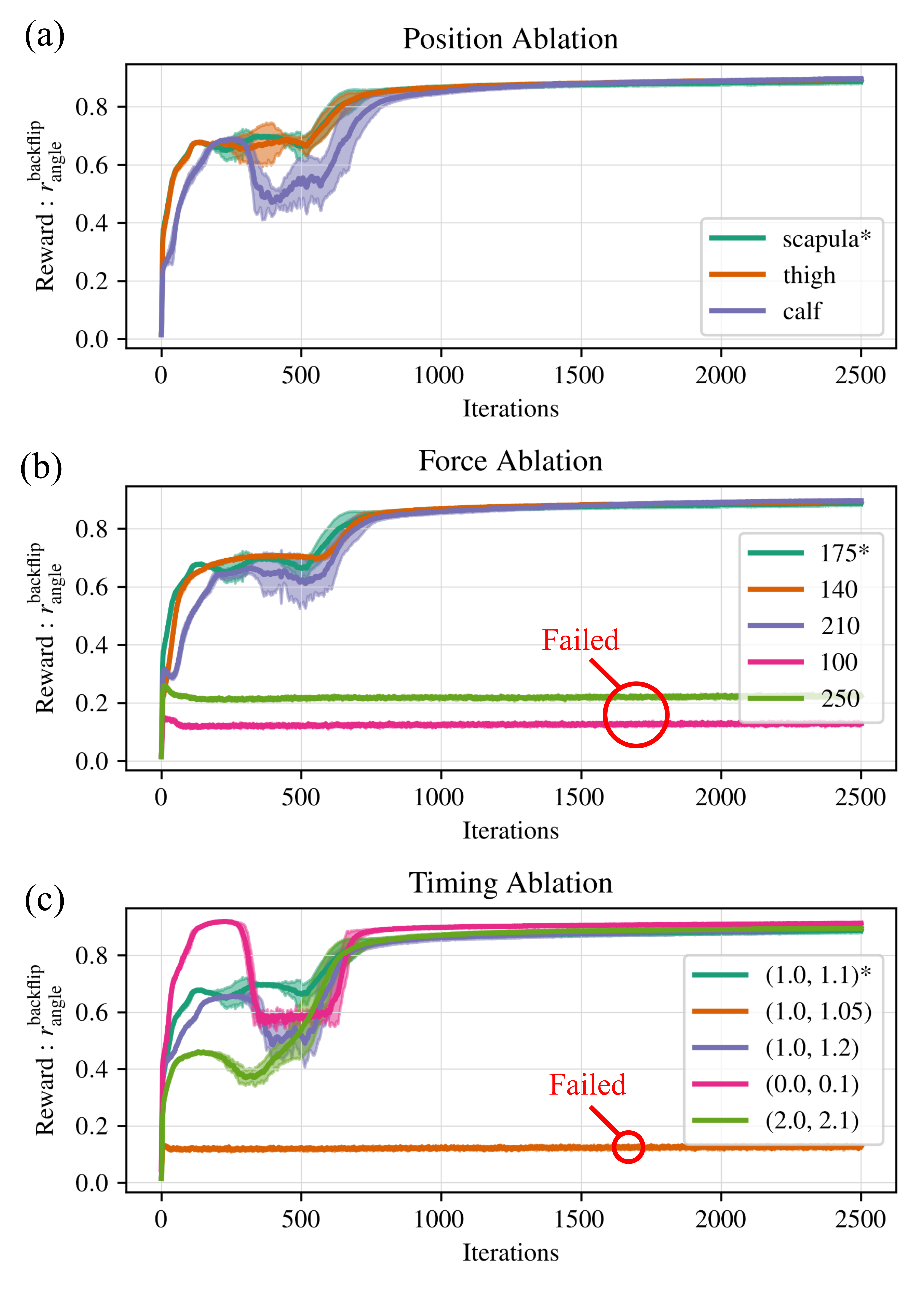}
	\vspace{-3ex}
	\caption{10. 補助力設計に関するアブレーション実験. (a) 作用点,(b) 力の大きさ,(c) 作用タイミングの影響. 広い条件範囲で学習が安定して進行する一方, 補助力が不適切な場合には学習が失敗することが確認できる.}
	\vspace{-3ex}
	\label{fig:assist-force-ablation}
\end{figure}

\subsection{価値関数推定の高速化に関する評価}\label{sec:experiment_value_function_compare}

外部補助によって Critic の学習が加速されるという仮説を検証するため,Jump タスクにおける価値関数推定を解析した.
\figref{fig:value_function_compare} は,学習の異なる段階において,成功動作に対応する価値関数の出力を比較したものである.

EFGCL を用いた場合,価値推定は 200 イテレーション時点ですでに最終的な分布に近い形へと収束した.
一方で,ベースラインでは同程度の精度に達するまでに 1{,}000 イテレーション以上を要し,推定値の分散も大きかった.
この結果は,学習初期に成功状態を経験することが,価値関数推定を大きく加速することを示している.
\begin{figure}[tbp]
\centering
	\includegraphics[width=0.95\columnwidth]{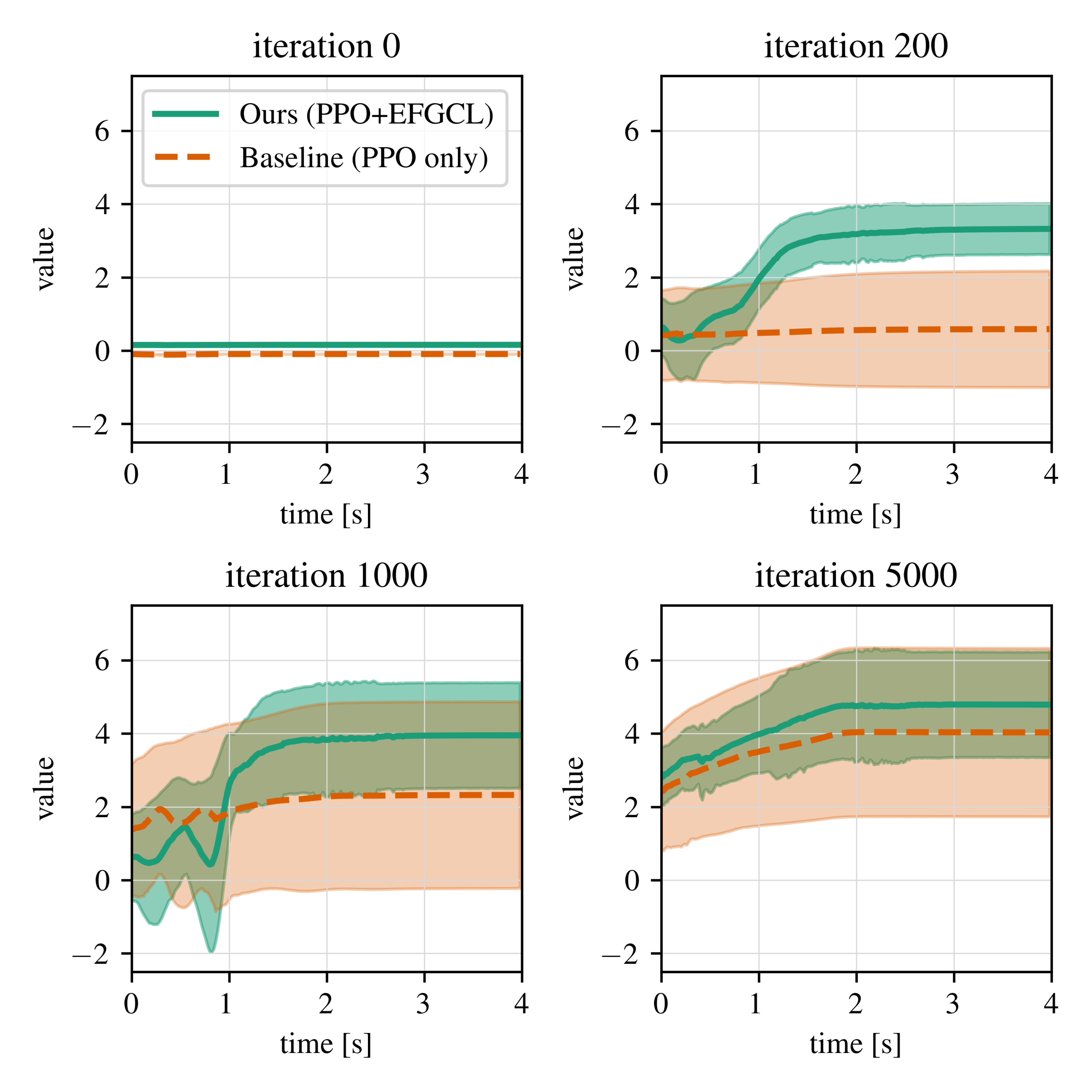}
	\vspace{-3ex}
	\caption{EFGCL の有無による Value Function 推定の比較. EFGCL を用いた場合,学習初期から最終的な価値推定に近い分布が得られ, 価値推定の高速化と安定化が確認できる.}
	\vspace{-3ex}
	\label{fig:value_fuction_compare}
\end{figure}
}
{
}

\section{Discussion}\label{sec:discussion}
\switchtext
{
\subsection{Efficacy and Robustness of Guided Exploration}
The experimental results demonstrate that EFGCL significantly stabilizes the learning of dynamic motion skills by accelerating value function estimation during the early training phase.
Unlike reward shaping or imitation learning, which rely on complex reward design or expert datasets, EFGCL guides exploration through direct physical assistance in the form of external forces.

Furthermore, the ablation study shows that the proposed method is highly robust to variations in assistive force design.
As long as the assistance roughly facilitates the target motion, learning can succeed without precise parameter tuning.
These results suggest that the principle of ``physically experiencing success'' provides a general and cost-effective strategy for overcoming exploration challenges in dynamic robotic reinforcement learning.

\subsection{Limitations and Future Work}
This study focuses on validating the principle of artificially enabling agents to experience successful motions.
Accordingly, the assistive forces were designed based on task-specific physical intuition.
While such heuristic designs are sufficient for the single-shot dynamic motion tasks considered in this work, the design burden may increase for more complex and continuous motions.

For continuous behaviors such as dancing, maintaining the overall motion structure often requires learning based on reference trajectories.
In such cases, using the proposed framework of external forces with sparse rewards alone may be insufficient, and combining it with trajectory-tracking reward designs or imitation learning is likely to be more effective.
Developing automatic optimization or generation methods for assistive forces that remain effective for complex target motions is an important direction for future work.
}
{
\subsection{Guided Exploration の有効性とロバスト性}
実験結果より,EFGCL は学習初期における価値関数推定を加速することで,動的運動スキルの学習を大きく安定化させることが確認された.
複雑な報酬設計や専門的なデータセットに依存する Reward Shaping や模倣学習とは異なり,EFGCL は外力による直接的な物理的補助を通じて探索を誘導する.

さらに,アブレーションスタディの結果から,提案手法は補助力設計のばらつきに対して高いロバスト性を有することが示された.
目標動作を大まかに支援する補助が与えられていれば,厳密なパラメータ調整を行わなくても学習は成立する.
この結果は,「物理的に成功を体験させる」という原理が,高い設計コストを伴うことなく探索困難性を克服するための,一般性のある戦略であることを示唆している.

\subsection{Limitations and Future Work}
本研究では, 成功動作を人工的に経験させるという原理の検証に主眼を置いたため, 補助力はタスクごとの物理的直感に基づいて設計した.
こうしたヒューリスティックな設計は, 本研究で扱ったような単発的な動的運動タスクに対しては十分であるが, より複雑で連続的な動作に対しては設計負担が増大する可能性がある.
ダンスなどの連続的な動作では, 動作全体の構造を維持するために, 参照軌道に基づく学習が適している場合が多い.
このような場合には, 本研究で提案した外力＋疎な報酬による枠組みを単独で用いるのではなく, 参照軌道追従を目的とした報酬設計や模倣学習と組み合わせることが有効である可能性が高い.
こうした複雑な目標動作に対しても有効な補助力の自動最適化・生成手法の開発は, 今後の重要な課題である.
}
{
}

\section{Conclusion}\label{sec:conclusion}
\switchtext
{
Inspired by spotting in gymnastics, we proposed External Force Guided Curriculum Learning (EFGCL), a reinforcement learning framework that guides exploration through decaying external forces.
Without relying on complex reward shaping or reference trajectories, EFGCL enables a quadrupedal robot to acquire dynamic whole-body motions, such as jumping and flipping, that are difficult for standard RL methods.

Through both simulation and real-robot experiments, we demonstrated successful sim-to-real transfer and showed that physical assistance accelerates value function estimation by allowing the agent to experience successful states early in training.
Although the current approach relies on heuristic force design, the results suggest that \emph{physical guidance} represents a promising and general paradigm for guided exploration, complementary to reward-based and imitation-based methods, in the learning of complex whole-body motions.
}
{
本研究では,体操競技における Spotting に着想を得て,外部補助力を段階的に減衰させながら探索を誘導する強化学習手法,External Force Guided Curriculum Learning(EFGCL)を提案した.
複雑な報酬設計や参照軌道に依存することなく,EFGCL はジャンプや宙返りといった,従来の強化学習手法では獲得が困難であった動的な全身運動を四脚ロボットに学習させることを可能にした.

シミュレーションおよび実機ロボットによる実験を通じて,提案手法がシミュレーションから実機への転移においても有効であること,ならびに外部補助によって学習初期に成功状態を経験させることが価値関数推定を加速することを確認した.
補助力の設計にはヒューリスティクスを用いているものの,本研究の結果は,Physical Guidance に基づく探索誘導が,報酬設計や模倣学習と相補的な,新たな Guided-RL のパラダイムとして,複雑な全身運動の学習において有効かつ汎用的な戦略であることを示唆している.
}
{
}

\appendices
\switchtext
{
\section{Reward Definitions}\label{app:reward_detail}

Table~\ref{tab:reward_terms} summarizes the reward terms shared across all tasks and their definitions.
Here, $P_{col}$ denotes the set of link indices used to detect collisions with the ground.
This set includes 14 links in total: the body, scapula, thigh, and calf, excluding the feet.
The indicator function $\delta^{\text{term}}_t$ takes the value of $1$ if the episode terminates due to the trunk contacting the ground at time $t$, and $0$ otherwise.

\subsection{Task-Specific Target Variables}
Table~\ref{tab:task_reward} presents the task-specific definitions of the target variable $x_t$,
the target value $x^{\text{target}}$, and the normalization coefficient $s_x$
used in the task progress reward $\rho^{\text{task}}_t$ shown in Table~\ref{tab:reward_terms}.

\begin{table}[htbp]
\centering
\caption{Summary of reward terms shared across all tasks.}
\label{tab:reward_terms}
\begin{tabular}{ll}
\toprule
Reward term & Definition \\
\midrule
Task progress
&
$\rho^{\text{task}}_t
=
\exp\!\left(
-\| x_t - x^{\text{target}} \|^2 / s_x
\right)$
\\
Standing
&
$\rho^{\text{stand}}_t
=
\exp\!\left(-\frac{\| h_t \|^2}{0.01}\right)
+
\exp\!\left(
	-\frac{\| \mathbf{q}_t - \mathbf{q}^{\text{stand}} \|^2}{0.25}
\right)$
\\
Angular regularization
&
$r^{\text{ang}}_t
=
-
\| (\omega^{\text{non-target}}_t)^2 \|$
\\
Collision penalty
&
$-1.0 \times \sum_{i\in P_{col}} (f_{i,z} > 0.1)$
\\
Termination penalty
&
$-100 \times \delta^{\text{term}}_t$
\\
Joint velocity penalty
&
$-5 \times 10^{-4} \, \| \dot{\mathbf{q}}_t \|^2$
\\
Joint acceleration penalty
&
$-1 \times 10^{-7} \, \| \ddot{\mathbf{q}}_t \|^2$
\\
\bottomrule
\end{tabular}
\end{table}
\begin{table}[htbp]
\centering
\caption{Task-specific instantiations of the task progress reward.}
\label{tab:task_reward}
\begin{tabular}{lccc}
\toprule
Task & Target variable $x_t$ & $x^{\text{target}}$ & $s_x$ \\
\midrule
Jump
& $h^{\text{max}}_t$
& $h^{\text{target}}$
& $0.01$
\\
Backflip
& $\theta^{\text{pitch}}_t$
& $2\pi$
& $\pi^2$
\\
Lateral-Flip
& $\theta^{\text{roll}}_t$
& $2\pi$
& $\pi^2$
\\
\bottomrule
\end{tabular}
\end{table}
}
{
\section{Reward Function Details}\label{app:reward_detail}
\subsection{Overview of Reward Terms}

Table~\ref{tab:reward_terms} に,
全タスクで共通に用いられる報酬項の一覧とその定義を示す.
なお$P_{col}$ は地面との衝突を検出するリンクのインデックス集合であり,
足先を除く body, scapula, thigh, calf の計14リンクが含まれる.
$\delta^{\text{term}}_t$ は時刻$t$において
胴体が地面に接触し episode が終了した場合に1,
それ以外は0となる指示関数である.

\begin{table}[t]
\centering
\caption{Summary of reward terms shared across all tasks.}
\label{tab:reward_terms}
\begin{tabular}{ll}
\toprule
Reward term & Definition \\
\midrule
Task progress
&
$\rho^{\text{task}}_t
=
\exp\!\left(
-\| x_t - x^{\text{target}} \|^2 / s_x
\right)$
\\
Standing
&
$\rho^{\text{stand}}_t
=
\exp\!\left(-\| h_t \| / 0.01\right)
+
\exp\!\left(
-\| \mathbf{q}_t - \mathbf{q}^{\text{stand}} \|^2 / 0.25
\right)$
\\
Angular regularization
&
$r^{\text{ang}}_t
=
-
\| (\omega^{\text{non-target}}_t)^2 \|$
\\
Collision penalty
&
$-1.0 \times \sum_{i\in P_{col}} (f_{i,z} > 0.1)$
\\
Termination penalty
&
$-100 \times \delta^{\text{term}}_t$
\\
Joint velocity penalty
&
$-5 \times 10^{-4} \, \| \dot{\mathbf{q}}_t \|^2$
\\
Joint acceleration penalty
&
$-1 \times 10^{-7} \, \| \ddot{\mathbf{q}}_t \|^2$
\\
\bottomrule
\end{tabular}
\end{table}

\subsection{Task-Specific Target Variables}
Table~\ref{tab:reward_terms} に示したタスク進捗報酬 $\rho^{\text{task}}_t$ における, 各タスク固有の目標変数 $x_t$, 目標値 $x^{\text{target}}$, 正規化係数 $s_x$ の具体的な定義をTable~\ref{tab:task_reward} に示す.

\begin{table}[t]
\centering
\caption{Task-specific instantiations of the task progress reward.}
\label{tab:task_reward}
\begin{tabular}{lccc}
\toprule
Task & Target variable $x_t$ & $x^{\text{target}}$ & $s_x$ \\
\midrule
Jump
& $h^{\text{max}}_t$
& $h^{\text{target}}$
& $0.01$
\\
Backflip
& $\theta^{\text{pitch}}_t$
& $2\pi$
& $\pi^2$
\\
Lateral-Flip
& $\theta^{\text{roll}}_t$
& $2\pi$
& $\pi^2$
\\
\bottomrule
\end{tabular}
\end{table}

\section{Deriviation of Assistive Force}\label{app:support_force}
$f_{jump}(l)$ は質量mの物体に$\Delta t$秒間上向きの力を加え, 頂点で高さ$l$に到達するような力である.

このような力を求めるため力$f$で質量$m$の物体を$\Delta t$秒間上向きに押したとき, 頂点に到達する時間を$t=T$とする.
時刻$0\le t \le T$の間に物体が移動した距離を$l$とすると
$T-\Delta t=\frac{(f-mg)\Delta t}{mg}$, $l = \frac{1}{2}(f-mg)\Delta t \ T$

この２つの式から$T$を消去すると$f$についての以下の二次方程式を得る.

$$ f^2 - mgf - \frac{lm^2g}{\Delta t^2} = 0 $$

これを解くことで$f_{jump}(l)$が以下のように求まる.
	$$f_{jump}(l) := \frac{mg}{2} \left(1+\sqrt{1+\frac{8l}{g \ {\Delta t}^2}}\right)$$
}
{
}

{
  \bibliographystyle{IEEEtran}
  \bibliography{bib}
}

\end{document}